\documentclass[10pt,twocolumn,letterpaper]{article}

\usepackage{wacv}              % To produce the CAMERA-READY version
\usepackage{tikz}        % protocol schematic in sec/fig1_protocol.tex
\usepackage{multirow}    % \multirow in results tables

\definecolor{wacvblue}{rgb}{0.21,0.49,0.74}
\usepackage[pagebackref,breaklinks,colorlinks,allcolors=wacvblue]{hyperref}

\def\wacvPaperID{46} % *** Enter the WACV Paper ID here
\def\confName{WACV}
\def\confYear{2027}

\title{What Does a Temporal Benchmark Score Measure? Decomposing Channel Use in Video VLM Evaluation}

\author{
Farrukh Rahman\\
Microsoft\\
Georgia Institute of Technology\\
}

\begin{document}

\maketitle

\begin{abstract}
A score on a temporal video question answering benchmark is meant to measure that a model has temporal understanding, but it conflates two questions. (1)~is the question even temporal, does it need several frames and their order? And (2)~when it does, does the model recover the order from the pixels, or read it off the positional encoding (RoPE)? Most of a temporal score answers neither, since a single frame and answer priors often carry it, and the field's validity checks, frame-shuffle sensitivity and multi-frame gain, speak only to (1). For (2) we score a reversal conflict, frames reversed while RoPE stays forward, two ways: label-free, as the accuracy lost against the forward labels, and against reversed labels wherever labels transform deterministically under time reversal. This separates a model that follows the reversed content from one merely disrupted. The two readings dissociate two similarly scoring models at both model scales. Qwen3-VL answers the reversed event it actually sees, reading the visual sequence. Molmo2 correctly reports a backwards video when positions agree, and flipping only the RoPE indices flips its prediction to the forward events, i.e. it reads positions. We call these regimes \emph{visual-sequence-dominant} and \emph{position-dominant}. Controls show reversed content is processable whenever the channels agree, and that both families share a bias toward describing events as running forward (arrow-of-time prior) that, being similar in strength, cannot by itself account for the dissociation. The distinction matters because the two channels fail on opposite inputs: two models with the same score are not interchangeable.
\end{abstract}

\section{Introduction}
\label{sec:intro}

A score on a temporal video benchmark is meant to certify that a model understands time. But ``temporal'' hides two questions. The task question: does a question even require watching motion across frames, and in order? The channel question: when it does, which input pathway does the model read the order from? We call these pathways \emph{channels}, the distinct routes by which temporal information enters the model. Prior video understanding work typically uses frame-shuffle sensitivity~\cite{shuffinvariance_sevilla2021only, buch2022revisitingvideovideolanguageunderstanding} and the accuracy a model gains from the full video, which answer only the task question.

Video VLMs receive temporal information through several channels: the pixels (motion across frames), in-context timestamps, and positional encoding (e.g. RoPE~\cite{rope}). A benchmark score aggregates all of them, with weights that depend on the input and the model. Shuffle sensitivity tells us the aggregate temporal signal is nonzero, but not the weighting over which channels carry it.

\begin{figure}[t]
\centering
\includegraphics[width=\linewidth]{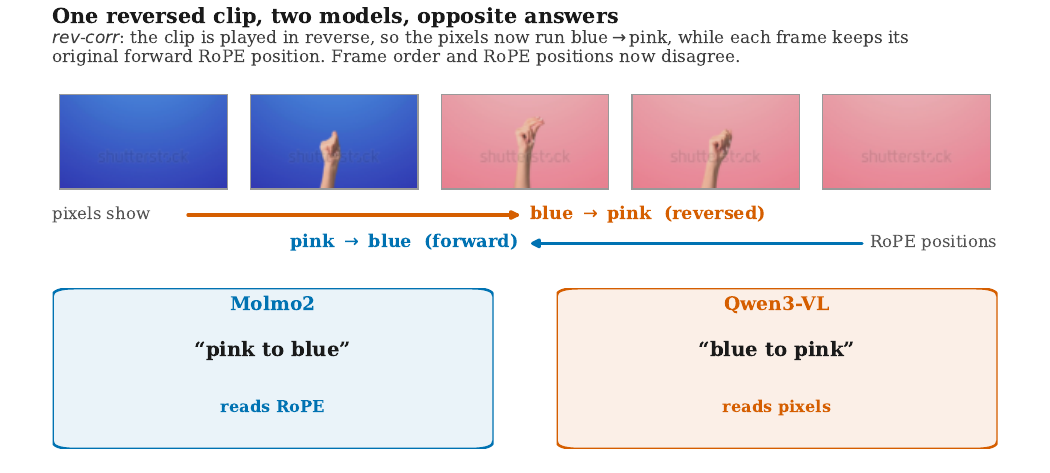}
\caption{
  In \texttt{rev-corr} the frames play in reverse, here a background changing color so the pixels run blue to pink, but each frame keeps its original forward RoPE index, so the pixels and the positions disagree. Molmo2 answers ``changing from pink to blue,'' reading order off positions. Qwen3-VL answers ``changing from blue to pink,'' following the visual sequence. The two models have a similar TempCompass score, yet the conflict separates their channel use (Sec.~\ref{sec:behavior:rope}).%
}
\label{fig:reversal-case}
\end{figure}

This gap matters, because two models with similar scores can resolve temporal order by opposite channels. Play a clip's frames in reverse but keep each frame's RoPE index at its forward value, so the pixels indicate one direction and the positions the other (Figure~\ref{fig:reversal-case}). Qwen3-VL answers the reversed event it actually sees, at $2.4\times$ chance against the reversed label. Molmo2 answers the forward event it was never shown, and shown the same backwards clip with positions agreeing it reports the backwards events correctly, so flipping only the RoPE indices flips its entire prediction (Table~\ref{tab:revgt}). Both models gain about $+0.20$ from the full video over a single frame, so a multi-frame-gain audit rates them identical. We call Molmo2 \emph{position-dominant} and Qwen3-VL \emph{visual-sequence-dominant}.

We make this concrete with an intervention protocol that perturbs the three channels independently and in conflict, across three video VLM families (Molmo2~\citep{molmo2}, Qwen3-VL~\citep{qwen3vl}, PerceptionLM~\citep{plm}) on two benchmarks (TempCompass~\citep{tempcompass}, TVBench~\citep{tvbench}). An information ladder bounds how much of a score is temporal and supplies a slice of \emph{motion-dependent} items, answered correctly from the full video but not from a single frame. The reversal conflict answers the channel question, and we score it against both the forward and the reversed label. The dissociation holds across both benchmarks and at two model scales, with controls: both families process reversed content when the channels agree, and the measured forward prior cannot by itself account for it (Section~\ref{sec:behavior:controls}).

Why is our conflict a clean probe of the positional channel? Because it damages nothing else. Every slot carries an intact frame, and only the binding between content and position changes. A probe analysis locates the consequence inside the network: the linearly decodable frame-order representation barely exists at the LM input and is constructed across the middle layers (Sec.~\ref{sec:behavior:patching}).

Our contributions:
\begin{itemize}[leftmargin=*]
    \item \textbf{A benchmark audit. A temporal score conflates 2 questions.} The aggregate score mixes whether a task needs motion with which channel resolves order when it does. We bound the task question with an information ladder (blank pixels, single frame, full video) and show that frame-shuffle sensitivity and multi-frame gain, the field's validity checks, answer only the first (Sec.~\ref{sec:behavior:audit}).
    \item \textbf{Two-label scoring for the reversal conflict.} Adopting the reversed-pixels, forward-RoPE intervention of~\citet{shi2025causality}, we contribute the scoring that makes it an evaluation diagnostic. Each prediction is scored against the forward label ($E$) and the reversed label ($F$), separating following-reversal from disruption, which a forward label drop alone cannot do. The label-free reversal-drop remains as a screen (Secs.~\ref{sec:methods:conflict},~\ref{sec:behavior:rope}).
    \item \textbf{A channel dissociation with a counterexample, at both scales.} Molmo2 reads order off RoPE, and RoPE correction alone flips which direction of events it reports, Qwen3-VL follows the visual sequence, and PerceptionLM sits between. Molmo2's position dominance is a counterexample to the causal-attention account of reversal sensitivity~\citep{shi2025causality} (Secs.~\ref{sec:behavior:rope},~\ref{sec:behavior:controls}).
    \item \textbf{Localization.} Activation patching localizes where shuffle damage is repairable, and a linear probe shows the frame-order code is constructed in the middle layers (Sec.~\ref{sec:behavior:patching}).
\end{itemize}

\section{Related Work}
\label{sec:related_work}

\textbf{Benchmark audits and what they measure.} A recurring finding in vision evaluation is that aggregate accuracy can measure a capability a model does not have, because models exploit shortcuts correlated with the label~\citep{geirhos2020shortcut, raji2021benchmark, geirhos2019texture}: many video questions are answerable from language, a single frame, or with the frames shuffled~\citep{shuffinvariance_sevilla2021only, buch2022revisitingvideovideolanguageunderstanding}. TempCompass~\citep{tempcompass} introduces diverse temporal aspects and paired reverse videos, TVBench~\citep{tvbench} rebuilds video VQA to resist single-frame and frame-shuffle shortcuts, and VBenchComp~\citep{vbenchcomp} sorts questions into language-answerable, shuffle-robust, and order-dependent. Their separators, single-frame sufficiency~\citep{buch2022revisitingvideovideolanguageunderstanding} and shuffle-invariance~\citep{shuffinvariance_sevilla2021only}, are black-box tests that identify \emph{which questions} need temporal order; we build on them and ask one question further: on those items, \emph{which channel} does a model use to resolve the order? The difference is mechanical: whole-clip reversal moves pixels and positions together and measures sensitivity, while the conflict we study reverses the pixels but holds the RoPE positions forward, isolating the channel. This requires access to model internals, so is limited to open models.

\textbf{The arrow of time.} \citet{xue2026arrow} report that large multimodal models largely fail to perceive the direction of time in videos. Our reversed-clip control measures the complementary bias on TempCompass. Shown reversed content as a perfectly normal video, six of seven models actively impose the forward direction, with $59$ to $75\%$ of their errors being exactly the forward-content answer against a random-error baseline of about $37\%$, while the weakest model shows no measurable forward bias (Sec~\ref{sec:behavior:controls}).

\textbf{Positional encodings for time.} Timestamp prompting~\citep{timechat} and absolute-time MRoPE~\citep{qwen3vl} \emph{introduce} positional temporal mechanisms into video VLMs. We show such encodings can \emph{override} the pixels during conflict, with reliance varying by architecture and question.

\textbf{Mechanistic localization.} Concurrent circuit-level work localizes temporal processing inside the VLM. CircuitProbe~\citep{circuitprobe} tracks the layerwise emergence of language-aligned object and action concepts, finds them decodable in mid-to-late layers, and reports models behaving closer to bag-of-frames evidence accumulation~\citep{buch2022revisitingvideovideolanguageunderstanding}. That work localizes where decoded features mature. We localize where the frame-order code is \emph{constructed}; it emerges across the middle layers and depends on positional encoding and is never built when the positional encoding is reversed (Sec~\ref{sec:behavior:patching}).

\textbf{Closest intervention.} \citet{shi2025causality} introduced the intervention pair this paper uses: their Reverse Order carries frames $N..1$ with each frame retaining its original position index, our \texttt{rev-corr}, and their Reverse PE is our mirror control. They find causal VideoLLMs largely insensitive to perturbing positional encodings yet sharply degraded under Reverse Order, attribute this to an order-sensitive causal-attention pathway, and contrast 3D MRoPE against 1D RoPE. We adopt their interventions and contribute two additions. First, a counterexample, Molmo2, bidirectional over vision tokens, does not degrade under the conflict but answers the forward event, so the causal-attention account is not universal, and the regime it misses is position-dominance. Second, the two-label scoring: their ground truth probability metric cannot distinguish a model that follows the reversed content from one disrupted by the conflict; scoring against both labels can (Sec~\ref{sec:behavior:rope}).

\definecolor{cFzero}{HTML}{0072B2}
\definecolor{cFone}{HTML}{E69F00}
\definecolor{cFtwo}{HTML}{009E73}
\definecolor{cFthree}{HTML}{D55E00}

\begin{figure*}[t]
\centering
\begin{tikzpicture}[
  chip/.style={rectangle, draw=black!30, line width=0.4pt,
    minimum width=0.40cm, minimum height=0.46cm, rounded corners=1.5pt,
    text=white, font=\bfseries\scriptsize, inner sep=0pt},
  tsc/.style={rectangle, draw=black!25, fill=black!4, line width=0.3pt,
    minimum width=0.40cm, minimum height=0.30cm, rounded corners=1pt,
    font=\scriptsize\ttfamily, text=black!75, inner sep=0pt},
  rpd/.style={font=\scriptsize\ttfamily, text=black!70},
  hdr/.style={font=\footnotesize\ttfamily, text=black!85, anchor=south},
  rowl/.style={font=\footnotesize\bfseries, anchor=east, text=black!80},
  rowg/.style={font=\scriptsize, anchor=east, text=black!55},
]

% ---- geometry ----------------------------------------------------------
\def\cw{1.84}      % column pitch
\def\xoff{0.35}    % left edge of first column
\def\cd{0.44}      % chip pitch within a column

% ---- highlight box behind the rev / rev-corr pair (columns 5,6) --------
\fill[black!6, rounded corners=3pt]
  ({\xoff+5*\cw+0.03},-1.86) rectangle ({\xoff+7*\cw-0.03},1.04);
\node[font=\scriptsize\itshape, text=black!60]
  at ({\xoff+6*\cw},-1.68) {only the RoPE row differs};

% ---- row labels with glosses ------------------------------------------
\node[rowl] at (0.05, 0.00) {Pixels};
\node[rowg] at (0.05,-0.26) {frame content};
\node[rowl] at (0.05,-0.64) {Timestamps};
\node[rowg] at (0.05,-0.90) {in-context text};
\node[rowl] at (0.05,-1.22) {RoPE};
\node[rowg] at (0.05,-1.48) {position indices};

% ---- condition columns -------------------------------------------------
% code / pixels (frame ids in presentation order) / timestamp digits / RoPE digits
\foreach \i/\code/\pix/\tss/\rp in {%
  0/base/{0,1,2,3}/{0,1,2,3}/{0,1,2,3},%
  1/shuf/{2,0,3,1}/{0,1,2,3}/{0,1,2,3},%
  2/shuf-corr/{2,0,3,1}/{0,1,2,3}/{2,0,3,1},%
  3/rop-shuf/{0,1,2,3}/{0,1,2,3}/{3,1,0,2},%
  4/conf-ts/{0,1,2,3}/{3,2,1,0}/{0,1,2,3},%
  5/rev/{3,2,1,0}/{0,1,2,3}/{0,1,2,3},%
  6/rev-corr/{3,2,1,0}/{0,1,2,3}/{3,2,1,0},%
  7/rev-mvts/{3,2,1,0}/{3,2,1,0}/{0,1,2,3}}{%
  \pgfmathsetmacro{\xc}{\xoff + (\i+0.5)*\cw}
  \node[hdr] at (\xc, 0.62) {\code};
  % pixels row
  \foreach \f [count=\k from 0] in \pix {%
    \pgfmathsetmacro{\xk}{\xc + (\k-1.5)*\cd}
    \pgfmathparse{int(\f)}%
    \ifcase\pgfmathresult
      \node[chip, fill=cFzero]  at (\xk, 0) {F0};\or
      \node[chip, fill=cFone]   at (\xk, 0) {F1};\or
      \node[chip, fill=cFtwo]   at (\xk, 0) {F2};\or
      \node[chip, fill=cFthree] at (\xk, 0) {F3};\fi
  }
  % timestamps row
  \foreach \t [count=\k from 0] in \tss {%
    \pgfmathsetmacro{\xk}{\xc + (\k-1.5)*\cd}
    \node[tsc] at (\xk, -0.64) {\t s};
  }
  % RoPE row
  \foreach \r [count=\k from 0] in \rp {%
    \pgfmathsetmacro{\xk}{\xc + (\k-1.5)*\cd}
    \node[rpd] at (\xk, -1.22) {\r};
  }
}

\end{tikzpicture}
\caption{%
The intervention protocol as a condition grid: each column is one
condition, each row one temporal channel. Pixels: the frames in
presentation order. Timestamps: the in-context time text preceding each
frame. RoPE: the position index each frame's vision tokens carry into
attention. When the RoPE digits ascend left to right, the positions
describe the presentation order, so the input is coherent even if reversed
(\texttt{rev} is a coherent backwards video). When they follow the frame
numbers instead, the positions encode the \emph{original} order and
disagree with the presentation: in \texttt{rev-corr} the pixels indicate
one temporal direction and the positions the opposite. In the highlighted
pair only the RoPE row changes, and Molmo2's reported direction of events
flips with it (Table~\ref{tab:revgt}). \texttt{-nts} variants drop the
timestamp row; \texttt{shuf-mvts} and \texttt{rev-mvts} move it with the
frames.%
}
\label{fig:protocol-3c}
\end{figure*}
\section{Methods}
\label{sec:methods}

\subsection{Models}
\label{sec:methods:models}

We study 3 video VLM families spanning 2 attention architectures and 2 positional encodings (1D vs. 3D RoPE).

\textbf{Qwen3-VL}~\citep{qwen3vl} (4B, 8B): Causal attention throughout. Vision tokens use 3D MRoPE with temporal, height, and width axes; the temporal index is replicated as the base for the H/W axes, so temporal ordering is encoded redundantly across all three axes. DeepStack~\cite{meng2024deepstackdeeplystackingvisual} injects intermediate ViT features into LLM layers 0--2. \\
\textbf{Molmo2}~\citep{molmo2} (4B, 8B, O-7B): 1D sequential RoPE with bidirectional attention for vision tokens (\texttt{token\_type\_ids}) and causal attention for text. Inter-frame information flow during prefill is constrained only by RoPE, not by sequential attention masking. \\
\textbf{PerceptionLM}~\citep{plm} (1B, 3B): 1D sequential RoPE, fully causal attention, no text timestamp format. \\
Behavioral interventions run on all 7 models, linear probes on Qwen3-VL and Molmo2.

\subsection{Benchmarks}
\label{sec:methods:benchmarks}

\textbf{TempCompass}~\citep{tempcompass}. 410 videos, $n=1580$ multiple choice questions across five aspects (action, attribute\_change, direction, order, speed). The audit (Sec~\ref{sec:behavior:audit}) motivates restricting per-aspect analysis to the three order-diagnostic aspects (attribute\_change, direction, order). \\
\textbf{TVBench}~\citep{tvbench}: 2205 multiple choice questions across nine categories; we use macro-averaged accuracy unless noted. Its temporal tasks are \emph{scene\_transition}, \emph{moving\_direction}, \emph{action\_sequence}, \emph{action\_localization}, and \emph{egocentric\_sequence}. \\
We use these two benchmarks because the designs are complementary to our question: TempCompass ships paired reverse ground truth, giving the paired reading directly, and TVBench is constructed to resist single-frame and shuffle shortcuts and supplies tasks whose labels transform deterministically under time reversal.

\textbf{Frames and sampling.}
Each model's preprocessing samples the frames it sees, so per-clip frame counts vary. TempCompass the median is 22 frames for Molmo2 (range 5--72) and 10 for Qwen3-VL (range 2--35). On TVBench the median is 36 for Molmo2 (cap 72) and 17 for Qwen3-VL, with the Qwen3-VL-8B TVBench runs sampling a constant 32 frames. All interventions operate on exactly these sampled frames, so the semantics of every condition, the RoPE spacing, and the DeepStack injections share the same per-clip frame count.

\subsection{Information Ladder}
\label{sec:methods:ladder}
To quantify how much of a benchmark score requires multiple frames, we evaluate each question at three information levels. 1.\ blank pixels with the timestamp text kept (\emph{blind}: answer priors and option leakage only), 2.\ a single repeated frame with its timestamp (\emph{single-frame}: adds static appearance), and 3.\ the full video. The \emph{multi-frame gain}, accuracy(full) minus accuracy(single-frame), isolates the accuracy that requires integrating across frames. For order we use the \emph{order-gain}, accuracy(full) $-$ accuracy(shuffled frames) (Sec.~\ref{sec:methods:interventions}), the shuffle-invariance criterion of TVBench~\citep{tvbench} and VBenchComp~\citep{vbenchcomp}. Single frame conditions keep the frame's timestamp, because a lone frame with its timestamp stripped is OOD for some models. An item is \emph{motion-dependent} for a model when it is answered correctly from the full video but not from a single frame. Sec.~\ref{sec:behavior:audit} validates this slice with an independent dose-response control and Sec.~\ref{sec:behavior:rope} uses it.

\subsection{Intervention Protocol}
\label{sec:methods:interventions}
We manipulate 3 input channels independently and in combination. Figure~\ref{fig:protocol-3c} draws the key conditions as a grid whose columns match Table~\ref{tab:conditions-main}; Supp.~\ref{app:conditions-full} has the complete reference. \\
\textbf{Visual order.} Frames are presented in one of: \emph{normal}, \emph{single-frame} (frame 0 repeated), \emph{shuffled}, \emph{reversed}, or \emph{blank} (zero-pixel frames). Post-Conv3d embedding shuffle yields equivalent results, ruling out Conv3d as an independent temporal channel (Supp.~\ref{app:patch-stage}). \\
\textbf{Timestamps.} Timestamps precede each frame's vision tokens. Manipulations: \emph{present}, \emph{absent}, \emph{mismatched} (shuffled independent of the visual shuffle), or \emph{reversed}. \\
\textbf{Positional encoding.} RoPE position indices for vision tokens are set to: \emph{native} (match input order), \emph{corrected} (follow original frame indices regardless of input order; under shuffle, slot $j$ receives the position of frame $\pi(j)$), or \emph{shuffled/reversed} (independently permuted or reversed). We additionally include correct-pixel, wrong-RoPE conditions (\texttt{rop-shuf}; \texttt{rop-rev} for Qwen only) to isolate RoPE effects with correct visual content. For Qwen3-VL, axis-specific MRoPE corrections recover per axis, DeepStack ablations show no independent temporal signal beyond RoPE, and RoPE-spacing warps show playback rate is not read from positional spacing (Supps.~\ref{app:axis-decomposition},~\ref{app:deepstack}).

\subsection{Conflict, Reversal, and Two-Label Scoring}
\label{sec:methods:conflict}

We construct conditions where channels indicate different orderings: \emph{forward visual + reversed timestamps} (\texttt{conf-ts}), \emph{consistent reversal of visual and timestamps} (\texttt{rev-mvts}), and \emph{reversed visual + corrected RoPE} (\texttt{rev-corr}). \emph{Reversal with native RoPE} (\texttt{rev}) is a coherent backwards video and serves as the no-conflict reference.

\textbf{Identifying the channel.} Under \texttt{rev-corr} the frames are reversed while the RoPE positions are held at the original forward order, so the pixels indicate one temporal direction and the positions the opposite. A model that reads order from positions answers the forward event; a model that reads order from the pixels answers the reversed event. A drop in forward label accuracy alone cannot distinguish the second behavior from generic disruption, so we score each prediction against both readings: $E$, agreement with the forward-content answer, and $F$, agreement with the reversed-content answer. Where no reversed label exists, the label-free \emph{reversal-drop}, accuracy(\texttt{base}) $-$ accuracy(\texttt{rev-corr}), serves as a screen (Sec.~\ref{sec:behavior:rope}).

\textbf{Reversed labels.} On TempCompass, 95 paired videos (\texttt{\{id\}} and \texttt{\{id\}\_reverse}) carry independently adjusted ground truth for \emph{attribute\_change} and \emph{direction}; the reversed label is matched by option text, restricted to informative pairs (original GT $\neq$ reversed GT). On TVBench, three tasks transform deterministically under time reversal: \emph{moving\_direction} (both motion axes flip, and the flipped option is always among the four candidates), \emph{scene\_transition} (``from A to B'' becomes ``from B to A''), and \emph{action\_sequence} (first and last swap, and both candidates are actions that occur in the clip by construction~\citep{tvbench}), so the reversed label is the other candidate. Items whose reversed label is not among the offered options are excluded rather than scored. The paired \texttt{\_reverse} videos under \texttt{rev-corr} instantiate the mirror conflict, forward pixels with reversed RoPE (Sec~\ref{sec:behavior:rope}, Supp.~\ref{app:dissociation}).

\textbf{Reversed-content control.} The 95 \texttt{\_reverse} clips are themselves valid forward videos, so their \texttt{base} accuracy against their own ground truth measures the intrinsic difficulty of reversed content with no intervention involved. Items are paired with the originals by (video, aspect, question stem). Ref Supp.~\ref{app:revclip}.

\textbf{Timestamp-sensitivity partition.}
To expose question-level timestamp use hidden by aggregate accuracy, we partition each model's questions into those answered correctly only when timestamps are present (\texttt{base} right, \texttt{base-nts} wrong) and those correct in both, then test whether reversing timestamps selectively affects the first set. This is a descriptive, model-specific partition, not a calibrated reliance estimate. Notation and full results in Supp.~\ref{app:subset-aspect}.

\textbf{Statistical conventions.}
Key shuffled conditions run with three permutation seeds Supp.~\ref{app:seeds}. A difference below $2$\,sd of the noisiest same-benchmark shuffled condition is treated as within permutation noise (about $2.5$ points). Two-label quantities carry paired bootstrap 95\% CIs (10000 resamples). Reversal conditions are read against reversed-content base accuracy (the reversed-content control), not original base. Decoding is greedy throughout.

\subsection{Activation Patching}
\label{sec:methods:patching}

We use residual-stream activation patching~\cite{patch, zhang2024bestpracticesactivationpatching} to localize \emph{where} clean-order content suffices to recover behavior after a shuffle. Activations at vision token positions, captured via a pre-hook at the input of each LM block on a \texttt{base} pass, are installed at the matching positions of a target pass (\texttt{shuf} unless noted) at the selected LM layers, leaving text tokens and RoPE indices untouched (positional encoding enters at each attention block's Q/K projections). Installation uses the original-frame correspondence, slot $\pi(j)$'s captured activation to slot $j$, so every patched token at every patched layer carries the activation it would have had in the unshuffled run.\\
\textbf{Patch extents and hooks.}
We test five extents, from \texttt{act\_layer0} (layer~0 only) to \texttt{act\_all}, including \texttt{act\_ds} (LM layers $\{0,1,2\}$, the DeepStack injection depth in Qwen3-VL); definitions and per-model layer boundaries in Supp.~\ref{app:patching-full}. Pre-hooks fire at the input of each LM block; in Qwen3-VL, DeepStack adds intermediate ViT features at vision-token positions \emph{inside} each of layers~0--2's forward passes, after the input pre-hook, so \texttt{act\_layer0} does not prevent re-injection of shuffled features at layers~1--2 and \texttt{act\_ds} is the depth at which all DeepStack injections have fired. Linear probes on the residual stream (Supp.~\ref{app:probe-detail}) provide the correlational counterpart used in Sec~\ref{sec:behavior:patching}.

\section{Results}
\label{sec:behavior}

We ask two questions of a temporal score: how much of it requires watching motion across frames in order (the task question), and, on the part that does, which channel a model uses to resolve temporal order (the channel question). Sec.~\ref{sec:behavior:audit} bounds the first and shows the standard diagnostics stop there. Secs~\ref{sec:behavior:rope} through~\ref{sec:behavior:patching} answer the second with the reversal conflict, its controls, and the localization.

\textbf{Reading effect sizes.}
Accuracies are decimals and a point is $0.01$. Chance is $0.315$ on TVBench (macro-averaged) and varies with the TempCompass aspect (2--4 options). The largest observed seed spread across shuffled conditions is sd $0.012$, both for micro-averaged scores and for the TVBench nine-task macro average (Supp.~\ref{app:seeds}). We treat a difference below $2$\,sd of the noisiest shuffled condition as within permutation noise ($2.5$ points). The rule applies to benchmark-level aggregates, paired reversal quantities carry bootstrap CIs and are read against reversed-content base accuracy (Sec.~\ref{sec:methods:conflict}).

\begin{table}[t]
\centering
\small
\caption{Core conditions used in main-text results. ``Native'' RoPE matches input order; ``corrected'' follows original frame indices. Full reference in Supp.~\ref{app:conditions-full}.}
\label{tab:conditions-main}
\begin{tabular}{llll}
\toprule
\textbf{Code} & \textbf{Pixels} & \textbf{Timestamps} & \textbf{RoPE} \\
\midrule
\texttt{base}      & original  & original        & native    \\
\texttt{base-nts}  & original  & absent          & native    \\
\texttt{shuf}      & shuffled  & at orig.\ slots & native    \\
\texttt{shuf-mvts} & shuffled  & moved w/ frames & native    \\
\texttt{shuf-corr} & shuffled  & at orig.\ slots & corrected \\
\texttt{conf-ts}   & original  & reversed        & native    \\
\texttt{rop-shuf}  & original  & original        & shuffled  \\
\texttt{rev}       & reversed  & at orig.\ slots & native    \\
\texttt{rev-corr}  & reversed  & at orig.\ slots & corrected \\
\texttt{rev-mvts}  & reversed  & moved w/ frames & native    \\
\bottomrule
\end{tabular}
\end{table}

\subsection{How Much of a Temporal Score Is Temporal?}
\label{sec:behavior:audit}

The information ladder (Sec.~\ref{sec:methods:ladder}) measures multi-frame dependence through the \emph{multi-frame gain} and order dependence through the \emph{order-gain}, the shuffle-invariance criterion of TVBench~\citep{tvbench} and VBenchComp~\citep{vbenchcomp} (Table~\ref{tab:audit}). Multi-frame gain measures dependence on multiple frames, not on their order, so a positive multi-frame gain can co-occur with order invariance.

Both families have comparable multi-frame gain, about $+0.20$ overall, so both need several frames. Order-gain is what separates the aspects: it concentrates on \emph{attribute\_change} and \emph{order}, is modest on \emph{direction}, and is near zero on \emph{action} and \emph{speed}. \emph{Action} is near ceiling but barely uses the pixels: its \emph{blind} accuracy is already about $0.5$ against chance $0.29$, so the score reflects single-frame recognition and answer-prior leakage. \emph{Speed} is the clean separator: a clear multi-frame gain with near-zero order-gain, so it needs several frames yet not their order, an aspect-level instance of VBenchComp's shuffle-robust category, consistent with Section~\ref{sec:behavior:subset}.

Order-gain also removes a confound: the \emph{order} aspect's single-frame baseline uses frame~0, which leaks which event comes first, while the shuffled baseline does not. A dose-response check confirms it further with a \emph{middle} frame $k{=}1$ baseline, accuracy on motion-dependent items climbs monotonically as frames are added while the static remainder stays flat from $k{=}1$ on (Supp.~\ref{app:doseresponse}). The motion-dependent subset is therefore validated with a baseline that does not share the frame-0 leak, and Sec~\ref{sec:behavior:rope} uses it.

\begin{table}[t]
\centering
\small
\caption{TempCompass per-aspect, averaged within family (Molmo2: 4B/8B/O-7B; Qwen3-VL: 4B/8B; per-scale values in Supp.~\ref{app:tempcompass}, Tab.~\ref{tab:audit-perscale}). \emph{Multi-frame gain} (full video $-$ single frame) measures dependence on multiple frames. \emph{Order-gain} (full video $-$ shuffled frames, shuffle invariance \cite{tvbench,vbenchcomp}) measures dependence on their order; shuffled accuracies are three-seed means (Supp.~\ref{app:seeds}). Speed needs frames but not order.}
\label{tab:audit}
\begin{tabular}{lcccc}
\toprule
 & \multicolumn{2}{c}{\textbf{Multi-frame gain}} & \multicolumn{2}{c}{\textbf{Order-gain}} \\
\cmidrule(lr){2-3}\cmidrule(lr){4-5}
\textbf{Aspect} & Molmo & Qwen & Molmo & Qwen \\
\midrule
attribute\_change & $+0.46$ & $+0.52$ & $+0.37$ & $+0.37$ \\
order             & $+0.29$ & $+0.21$ & $+0.33$ & $+0.30$ \\
direction         & $+0.17$ & $+0.19$ & $+0.14$ & $+0.12$ \\
speed             & $+0.11$ & $+0.11$ & $+0.05$ & $+0.01$ \\
action            & $+0.09$ & $+0.05$ & $+0.01$ & $+0.01$ \\
\midrule
Overall           & $+0.22$ & $+0.21$ & $+0.17$ & $+0.15$ \\
\bottomrule
\end{tabular}
\end{table}

\textbf{Shuffle sensitivity cannot identify the channel.}
Disrupting RoPE alone while leaving the pixels intact (\texttt{rop-shuf}) costs Molmo nearly as much as a full pixel shuffle: with one permutation shared across both conditions per video, the stimulus-matched gap is at most $3.4$ points at any scale, against shuffle drops of $18$ to $29$ (Supp.~\ref{app:seeds}, Table~\ref{tab:tvbench-molmo}). The same manipulation costs Qwen only $3$ to $6$ points, consistent with its 3D MRoPE encoding order redundantly across axes (Supp.~\ref{app:axis-decomposition}). Both models thus look alike to the standard diagnostics yet resolve order by opposite channels, which the next section measures directly.

\subsection{The Channel Dissociation, Scored Against Reversed Labels}
\label{sec:behavior:rope}

Under \texttt{rev-corr} the pixels indicate one temporal direction and the positions the opposite, and we score each prediction against both readings (Sec.~\ref{sec:methods:conflict}): $E$, the forward-content answer, and $F$, the reversed-content answer, on the three TVBench tasks whose labels flip deterministically under reversal. Alongside \texttt{rev-corr} we report \texttt{rev}, reversed pixels with native RoPE: a coherent backwards video and the no-conflict reference.

\begin{table}[t]
\centering
\small
\setlength{\tabcolsep}{3.2pt}
\caption{
Two-label reversal test on TVBench.
$F$ scores the reversed event and $E$ the forward event.
\texttt{rev} is a coherent backwards video; \texttt{rev-corr}
(\texttt{r-c}) uses the same reversed frames with forward RoPE,
so the two conditions differ only in position indices.
Under \texttt{r-c}, high $F$ indicates following visual order,
while high $E$ indicates following positional order.
\emph{moving\_direction}: $n{=}232$, chance $0.25$;
\emph{scene\_transition}: $n{=}184$, chance $0.5$.
Full results and the \emph{moving\_direction} asymmetry are in
Supp.~\ref{app:revgt}.
}
\label{tab:revgt}
\resizebox{\columnwidth}{!}{%
\begin{tabular}{lcccc@{\hskip 8pt}cccc}
\toprule
 & \multicolumn{4}{c}{\emph{moving\_direction}}
 & \multicolumn{4}{c}{\emph{scene\_transition}} \\
\cmidrule(lr){2-5}\cmidrule(lr){6-9}
\textbf{Model}
& \texttt{rev} $F$
& \texttt{r-c} $F$
& \texttt{r-c} $E$
& \texttt{base} $E$
& \texttt{rev} $F$
& \texttt{r-c} $F$
& \texttt{r-c} $E$
& \texttt{base} $E$ \\
\midrule
Qwen 4B  & 0.63 & 0.61 & 0.06 & 0.47 & 0.78 & 0.69 & 0.31 & 0.87 \\
Qwen 8B  & 0.79 & 0.65 & 0.05 & 0.38 & 0.77 & 0.69 & 0.32 & 0.87 \\
Molmo 4B & 0.65 & 0.05 & 0.62 & 0.70 & 0.78 & 0.13 & 0.88 & 0.89 \\
Molmo 8B & 0.60 & 0.08 & 0.67 & 0.69 & 0.78 & 0.15 & 0.85 & 0.85 \\
\bottomrule
\end{tabular}}
\end{table}

\textbf{Qwen follows visual order.}
Under the conflict Qwen answers the reversed event at $0.61$/$0.65$ on \emph{moving\_direction} (chance $0.25$), and at $0.69$ on \emph{scene\_transition} (chance $0.5$), nearly matching its accuracy on the coherent backwards video, with a distractor rate no higher than under \texttt{base}. It is not disrupted by the conflict; it tracks the reversed visuals despite conflicting positions. The conflict does cost more at 8B than 4B (\texttt{rev} minus \texttt{rev-corr} $F$: $0.02$ vs $0.14$).

\textbf{RoPE alone flips Molmo's report.}
On the coherent backwards video Molmo correctly reports the reversed events ($F$ $0.60$ to $0.78$), so it is not insensitive to reversal. Under \texttt{rev-corr} its report flips to the forward events at near-\texttt{base} accuracy ($E$ $0.62$ to $0.88$). Between those two conditions nothing changes except the RoPE indices. Molmo reads temporal order off positions, and when positions and pixels disagree, the positions win. We call Molmo \emph{position-dominant} and Qwen \emph{visual-sequence-dominant}.

\textbf{Scope: \emph{action\_sequence}.}
The task is uninformative for Qwen, which sits near the two-option chance even on the coherent backwards video ($F$ $0.57$/$0.55$ under \texttt{rev}): reversed everyday actions are hard for it regardless of conflict, consistent with the arrow-of-time findings of~\citet{xue2026arrow}. Molmo's override remains clean there (\texttt{rev-corr} $E{=}0.82$, equal to \texttt{base}; \texttt{rev} $F{=}0.62$/$0.64$).

\textbf{The label-free reversal-drop is the screen.}
For benchmarks or tasks without reversed labels we read the \emph{reversal-drop}, accuracy(\texttt{base}) $-$ accuracy(\texttt{rev-corr}) against the forward labels (per-task values with paired bootstrap CIs in Supp.~\ref{app:tvbench}, Table~\ref{tab:dissoc-tvb}). Molmo's drops stay at or below $+0.08$ on the TVBench temporal tasks at both scales, with one clear exception (\emph{egocentric\_sequence} at 8B, $+0.14$), while Qwen drops $0.24$ to $0.55$ on every temporal task, every CI excluding zero. The cost is also specific: outside that one exception Molmo sits near zero on all nine tasks, while Qwen's cost concentrates on the temporal five (its largest non-temporal drop, $+0.18$, is well below its smallest temporal one, $+0.24$). Wherever the screen could be checked against reversed labels, the paired reading confirmed it. The drop alone remains a screen, not a classifier.

\textbf{TempCompass: paired ground truth, and what the all-pairs metric hides.}
TempCompass's paired \texttt{\_reverse} annotations give the same two-label scoring on \emph{attribute\_change} (Table~\ref{tab:reversal-tc}). Molmo shows a near-total positional override at every scale ($F{-}E$ $-0.82$ to $-0.96$ over all informative pairs). Qwen's all-pairs value is also negative ($-0.23$/$-0.14$), but for a reason unrelated to positions: many items are answerable from a single frame, and reversal does not change what any one frame shows, so a statically answering model returns the forward answer whatever its channel. Such items drag $F{-}E$ negative regardless of position use (Supp.~\ref{app:dissociation}). The motion-dependent slice removes them, and the numbers there agree with TVBench: a statistical zero at 4B ($+0.01$, CI $[-0.21,+0.23]$, $n{=}71$) and positive at 8B ($+0.24$, CI $[+0.01,+0.45]$, $n{=}76$), against Molmo's $-0.98$ to $-1.00$.

A held-out check addresses outcome selection: slices defined by one family and evaluated on the other reproduce the pattern, with the held-out 8B CI excluding zero (Supp.~\ref{app:heldout}). The mirror conflict (forward pixels, reversed RoPE) also flips the family pattern, confirming the dissociation in the opposite direction (Table~\ref{tab:reversal-tc}, \emph{mirror}).The second paired aspect, \emph{direction}, is less diagnostic: its order-gain is modest (Table~\ref{tab:audit}), and Qwen's accuracy is already low on coherent reversed direction clips. It therefore tests mainly the positional side: Molmo's override persists, while Qwen follows the forward-time prior.  (Sec.~\ref{sec:behavior:controls} Supp.~\ref{app:dissociation}).

\begin{table}[t]
\centering
\small
\caption{Channel dissociation on TempCompass \emph{attribute\_change}. 
\texttt{rev-corr} uses reversed pixels with forward RoPE; \emph{mirror} uses forward pixels with reversed RoPE. 
We report forward-label accuracy under \texttt{base}, \texttt{rev-corr}, and \emph{mirror}, and $F{-}E$ over all $n{=}141$ informative pairs and the motion-dependent subset ($n$ in parentheses). 
$F{-}E<0$ indicates the forward event (position-dominant); $F{-}E>0$ indicates the reversed event (visual-sequence-dominant). 
Paired bootstrap 95\% CIs. Per-aspect $E/F$, the static remainder, and \emph{direction} are in Supp.~\ref{app:dissociation}.}
\label{tab:reversal-tc}
\resizebox{\columnwidth}{!}{%
\begin{tabular}{lccccc}
\toprule
\textbf{Model} & \texttt{base} & \texttt{rev-corr} & \emph{mirror} & $F\!-\!E$ [CI] & $F\!-\!E$ m-dep [CI] ($n$) \\
\midrule
Molmo2 4B   & 0.965 & 0.965 & 0.312 & $-0.957$ $[-0.99,-0.91]$ & $-1.000$ $[-1.00,-1.00]$ (67) \\
Molmo2 8B   & 0.972 & 0.965 & 0.368 & $-0.950$ $[-0.99,-0.90]$ & $-0.983$ $[-1.00,-0.95]$ (59) \\
Molmo2 O-7B & 0.965 & 0.882 & 0.458 & $-0.816$ $[-0.90,-0.72]$ & $-0.766$ $[-0.90,-0.62]$ (77) \\
\midrule
Qwen3-VL 4B & 0.944 & 0.576 & 0.868 & $-0.234$ $[-0.39,-0.09]$ & $+0.014$ $[-0.21,+0.23]$ (71) \\
Qwen3-VL 8B & 0.972 & 0.549 & 0.896 & $-0.135$ $[-0.30,+0.03]$ & $+0.237$ $[+0.01,+0.45]$ (76) \\
\bottomrule
\end{tabular}}
\end{table}

TVBench establishes the dissociation at both scales. TempCompass independently confirms Molmo's positional override and the mirror pattern; Qwen's motion-dependent readout is positive at 8B and statistically null at 4B. On TempCompass the two models match in both score and multi-frame gain, so the standard reports see no difference at all, and on neither benchmark does any aggregate metric reveal which channel a model reads.

\subsection{Controls: Reversed Content Is Processable}
\label{sec:behavior:controls}

Two questions for Section~\ref{sec:behavior:rope} we must consider: that reversed videos are simply too far out of distribution to process, and that the drops reflect a generic difficulty of reversed content rather than the encoding.

\textbf{Consistent reversal.}
Under \texttt{rev-mvts}, frames and timestamps reverse together so every channel tells the same backwards story. Both families then track the reversed content well at both scales ($F$ $0.53$ to $0.80$ at 8B across the three flip tasks, with one 4B exception, Qwen \emph{moving\_direction}; Supp.~\ref{app:revgt}). Reversed content is processable whenever the channels agree, so the \texttt{rev-corr} effects are effects of the conflict, not of reversal per se. \\
\textbf{Shared forward prior.}
TempCompass ships 95 clips that are themselves reversed videos, viewed under completely normal conditions (\texttt{base}, no intervention); Table~\ref{tab:prior} gives the control. Reversed content is genuinely harder (CI excludes zero, Supp.~\ref{app:revclip}), and in $6/7$ models the errors are not random but the forward content answer, at about twice the random error rate. E.g. shown a clip of a demolition that is really a reversed construction time-lapse, the models say ``being constructed,'' a forward arrow of time imposed on backwards content, quantifying on this benchmark the bias reported by~\citet{xue2026arrow}. The exception is PLM 1B, the weakest model, whose errors sit at the random baseline.

The cost is similar in size across families; and because the dissociation contrasts (\texttt{rev} vs \texttt{rev-corr}) hold the reversed content fixed and vary the encoding, the shared prior cannot by itself explain the divergence between them. It also sets the floor as the prior alone yields forward answers on $30\%$ of reversed clips, while Molmo under \texttt{rev-corr} answers forward at $0.97$, and PLM 1B, with no measurable prior, at $0.72$. I.e. the override is positional, not semantic. 

\begin{table}[t]
\centering
\small
\setlength{\tabcolsep}{4pt}
\caption{Reversed-content control (TempCompass, 95 \texttt{\_reverse} clips under \texttt{base}): accuracy on the paired originals and on the reversed clips, the loss $\Delta$ (paired bootstrap CIs all exclude zero), the fraction of reversed-clip errors that are the forward-content answer (\emph{fwd-err}) against the random-error baseline (\emph{rand}), and, for comparison, the forward-answer rate under the \texttt{rev-corr} conflict on \emph{attribute\_change} informative pairs (\emph{r-c fwd}). Per-aspect values and CIs in Supp.~\ref{app:revclip}.}
\label{tab:prior}
\begin{tabular}{lcccccc}
\toprule
\textbf{Model} & orig & rev & $\Delta$ & fwd-err & rand & r-c fwd \\
\midrule
Molmo2 4B   & 0.79 & 0.56 & $-0.24$ & 0.73 & 0.37 & 0.97 \\
Molmo2 8B   & 0.79 & 0.57 & $-0.21$ & 0.73 & 0.37 & 0.97 \\
Molmo2 O-7B & 0.78 & 0.56 & $-0.22$ & 0.68 & 0.37 & 0.89 \\
Qwen3-VL 4B & 0.80 & 0.47 & $-0.33$ & 0.75 & 0.37 & 0.58 \\
Qwen3-VL 8B & 0.79 & 0.56 & $-0.23$ & 0.75 & 0.37 & 0.55 \\
PLM 1B      & 0.67 & 0.46 & $-0.22$ & 0.31 & 0.38 & 0.72 \\
PLM 3B      & 0.73 & 0.55 & $-0.19$ & 0.59 & 0.38 & 0.71 \\
\bottomrule
\end{tabular}
\end{table}

\subsection{Localization: Where Damage Can Be Repaired}
\label{sec:behavior:patching}

Activation patching localizes where shuffle-induced errors can be repaired. It cannot adjudicate the pixel-versus-position conflict because \texttt{rev-corr} preserves frame content and changes only its binding to position, while RoPE is re-applied inside each attention block rather than stored in the activations. The dissociation is therefore supported by the behavioral controls in Secs.~\ref{sec:behavior:rope} and~\ref{sec:behavior:controls}.\\
\textbf{Recovery tracks architecture.}
Layer-0 patching fully restores every model except Qwen3-VL, whose residual closes at \texttt{act\_ds}, once all DeepStack injections have fired (Sec.~\ref{sec:methods:patching}). Input space RoPE correction (\texttt{shuf-corr}) instead recovers only the bidirectional family. Under causal attention, K/V vectors commit to the shuffled context during prefill, and later correction cannot re-organize them (per-model data and TVBench replication in Supps.~\ref{app:patching-full},~\ref{app:tvbench}).\\
\textbf{Linear probing frame order.}
A linear probe for frame order, trained on clean-run layer inputs, decodes order at only $0.27$ entering layer~3 and $0.84$ by layer~20 (Qwen3-VL-4B), showing that frame order becomes increasingly linearly decodable across the middle layers. Under reversed RoPE, this increase does not occur, and installing a clean residual at layer~3 recovers nothing (behavioral delta vs. the reversed-RoPE run: $+0.000$/$+0.006$; Supp.~\ref{app:probe-detail}). Thus, the decodable order signal emerges downstream of the positional signal rather than being present in early representations. The same layerwise emergence appears in all five probed models; under shuffled positions, it is absent in Molmo (Supp.~\ref{app:probe-detail}, Table~\ref{tab:probe-layers}).

\section{Discussion}
\label{sec:discussion}

\subsection{Where the Order Comes From}
\label{sec:discussion:channels}

Our results show how temporal order enters these models. On a reversed input, 3 sources can set a model's reported direction: a semantic prior toward the canonical direction of events (reversed-content control) and the positional encoding (isolated by \texttt{rev-corr}) both push toward the forward events; the visual sequence (measured by \texttt{rev}) pushes toward the reversed ones. The protocol separates all 3.

Molmo2 is \emph{position-dominant}: RoPE correction alone flips its report of a backwards video to the forward events at near-base accuracy, at both scales. It is neither order-invariant, since shuffling still hurts it, nor reversal-insensitive, since it tracks reversed content when the channels agree. Instead, it reads order from position indices rather than visual sequence order, an ordered, position-indexed counterpart to the bag-of-frames behavior reported in prior video-language analyses~\citep{buch2022revisitingvideovideolanguageunderstanding,circuitprobe}.

Qwen3-VL is \emph{visual-sequence-dominant}. Under the conflict it answers the reversed event it actually sees, at $2.4\times$ chance and near its no-conflict accuracy. Its residual positional pull appears only on statically answerable items, is a statistical zero on the motion-dependent slice at 4B, and reverses sign at 8B. The two scale observations measure different things: the conflict costs 8B more reversed-content accuracy (Table~\ref{tab:revgt}), yet its remaining answers favor the content more decisively (Table~\ref{tab:reversal-tc}). Both readings leave 8B visual-sequence-dominant.

The distinction is not binary, but continuous and the weighting is a function of the question. PerceptionLM fills the middle with 1D positional encoding like Molmo2 but causal attention like Qwen3 ($F{-}E$ $-0.53$, $-0.49$ Supp.~\ref{app:plm}). The ordering Molmo2 $>$ PLM $>$ Qwen3 is consistent with both 1D positional encoding and bidirectional attention pushing a model toward reading order off positions, so neither factor alone accounts for this. This is viewed as consistency rather than attribution as the families differ in architecture and training data/objectives. Regardless, the point for evaluation stands as Molmo2 and Qwen3-VL look identical to a multi-frame-gain and to aggregate score yet the channel decomposition separates them.

\label{sec:behavior:subset}%
Timestamps are the clearest case of question dependent weighting: their two natural measures dissociate. Removing timestamps outright (\texttt{base-nts} vs.\ \texttt{base}) costs Molmo more than Qwen at every scale, and the cost grows with scale (Supp.~\ref{app:tempcompass}). Yet on the subset a model answers only when timestamps are present, \emph{reversing} them is decisive, and that penalty is smaller for Molmo and shrinks with scale, while on the rest it changes almost nothing (ref Supp.~\ref{app:subset-aspect}). Molmo uses timestamps on more questions but is less misled when they conflict with the pixels; no single number describes either model. RoPE order is carried in its continuous spacing too, not only the frame permutation, and playback rate is not read from positions at all (Supp.~\ref{app:axis-decomposition}).

The dissociation has a mechanistic counterpart in Sec~\ref{sec:behavior:patching}. The frame-order representations are constructed mid network by position driven attention rather than stored in early activations, which is why reversing RoPE with correct pixels reproduces the order cost and cannot be undone by installing correct activations at depth (Supp.~\ref{app:rope-depth}). \texttt{rev-corr} is a clean probe for the same reason, it perturbs only the coordinates the construction runs on.
\subsection{Limitations}
\label{sec:limitations}
The reversal conflict is an OOD diagnostic and does not estimate natural failure rates. All models are sub-10B, while the behavioral observations and controls replicate at both scales across both families, larger models could shift this balance; the protocol transfers unchanged to any open model. PerceptionLM rules out attention type or positional encoding alone as the cause, but the bidirectional result rests on Molmo2 and the families differ in training. All reversal-based diagnostics, ours included, operate on top of the shared forward prior of Sec~\ref{sec:behavior:controls}, so absolute accuracy on reversed content understates competence. Deterministic-flip tasks screen the channel only where a model can track reversed content at all, which \emph{action\_sequence} shows for Qwen3-VL. The timestamp partition is model-specific, and the experiments cover two short-video VQA benchmarks.

\subsection{Recommendations for Temporal Evaluation}
\label{sec:discussion:forward}

Overall recommendation is to separate the questions a single temporal score collapses. Whether a task needs multiple frames, whether it needs their order, and which channel resolves that order. Multi-frame gain addresses the first, order-gain the second, and the reversal protocol the third.

\begin{itemize}[leftmargin=*]
  \item \textbf{Report the decomposition, not aggregate accuracy.} A positive multi-frame gain does not imply order matters: \emph{speed} needs frames yet is order-invariant (Sec~\ref{sec:behavior:audit}).
  \item \textbf{Use the reversal conflict as a screen, paired with its controls.} A drop alone cannot separate following-reversal from disruption, and a near-zero drop cannot separate position use from insensitivity. Run \texttt{rev} alongside: a model that tracks reversed content without conflict and holds accuracy under conflict is reading positions.
  \item \textbf{Score against reversed labels wherever they exist, and design for them.} Tasks whose labels transform deterministically under time reversal (symmetric motion, orderings whose candidates occur in the clip) make the channel question answerable with no new annotation. A channel-probe item needs the deterministic flip, genuine order dependence, and time-symmetric content: a flippable label over arrow-of-time-violating content measures the shared prior instead (Secs.~\ref{sec:behavior:rope},~\ref{sec:behavior:controls}). Benchmark authors can build this in the way TVBench built in shuffle resistance.
\item \textbf{Do not infer channel use from aggregate accuracy.} Similar scores do not imply that two models resolve temporal order through the same channel. Compare reversal conditions to the reversed-content baseline (Sec.~\ref{sec:methods:conflict}).
\end{itemize}
\subsection{Conclusion}
\label{sec:discussion:conclusion}

Temporal scores conflate multi-frame dependence, order dependence, and the channel that resolves order. Across two benchmarks and two scales, similarly scoring Molmo2 and Qwen3-VL sit at opposite ends of that third axis: one reads positions, the other the visual sequence, and the order representation itself emerges in the middle layers. 2 models with the same temporal score are therefore not interchangeable, because the two channels fail on opposite inputs.

{
    \small
    \bibliographystyle{ieeenat_fullname}
    \bibliography{main}
}

\appendix
% ============================================================
% Supplementary Material
% ============================================================

\section*{Supplemental/Appendix}

% ------------------------------------------------------------
\section{Reproducibility}
\label{app:reproducibility}
% ------------------------------------------------------------

\paragraph{Models.} We evaluate seven models from three families, using their official released checkpoints: Qwen3-VL~\citep{qwen3vl} (4B, 8B), Molmo2~\citep{molmo2} (4B, 8B, O-7B), and PerceptionLM~\citep{plm} (1B, 3B). Behavioral interventions run on all seven models; linear probes on Qwen3-VL and Molmo2 only.

\paragraph{Benchmarks.} TempCompass~\citep{tempcompass} (410 videos; $n=1580$ questions, $1564$ on Molmo2 O-7B) and TVBench~\citep{tvbench} ($n=2205$ questions, nine categories), both multiple-choice. Per-aspect channel analysis uses attribute\_change, direction, and order; action and speed are excluded as non-diagnostic of order (Section~\ref{sec:behavior:audit}), while the audit (Table~\ref{tab:audit}) reports all five.

\paragraph{Interventions and the paired-GT split.} All conditions are defined in Table~\ref{tab:conditions-main} (main text) and Table~\ref{tab:conditions-full} (supplementary). The paired-ground-truth reversal split comprises 95 paired videos (144 attribute\_change and 157 direction pairs) and is reconstructed from TempCompass by the relabeling procedure of Section~\ref{sec:methods:conflict}.

\paragraph{Compute, Seeds and statistics.} Single-seed runs use seed~42. Tabulated shuffled conditions are single-run unless a caption says otherwise, with three-seed spreads in Supp.~\ref{app:seeds}; permutation seeds affect only shuffles. Paired bootstrap 95\% CIs use 10{,}000 resamples (percentile method) throughout, with one exception: the timestamp-partition CIs of Supp.~\ref{app:subset-aspect} use 1{,}000, stated there. All experiments ran on a single consumer NVIDIA RTX 4090 GPU.

\paragraph{Artifacts.} The paired-GT split is reconstructible from TempCompass via the outlined procedure.

% ------------------------------------------------------------
\section{Full Activation Patching Profiles}
\label{app:patching-full}
% ------------------------------------------------------------

Table~\ref{tab:mechanism-summary} summarizes the localization profile
across the three families that Section~\ref{sec:behavior:patching}
describes; Table~\ref{tab:patching} and Appendix~\ref{app:plm} give the
full per-model profiles.

\begin{table}[h]
\centering
\small
\caption{Localization profile across 3 model families under \emph{shuffle}-targeted patching. Decisive architectural differences are LM attention type and DeepStack. \texttt{act\_ds} patches exactly LM layers~\{0,1,2\} (DeepStack injection depth).}
\label{tab:mechanism-summary}
\resizebox{\columnwidth}{!}{%
\begin{tabular}{lcccccc}
\toprule
\textbf{Model} & \textbf{LM attn} & \textbf{DeepStack} & \textbf{shuf-corr gap} & \textbf{act\_layer0 gap} & \textbf{act\_ds gap} & \textbf{act\_early gap} \\
\midrule
Molmo2 4B   & Bidir. & No  & $-0.001$ & $\phantom{-}0.000$ & ---              & $\phantom{-}0.000$ \\
Molmo2 8B   & Bidir. & No  & $-0.004$ & $\phantom{-}0.000$ & ---              & $\phantom{-}0.000$ \\
Molmo2 O-7B & Bidir. & No  & $-0.051$ & $\phantom{-}0.000$ & ---              & $\phantom{-}0.000$ \\
\midrule
Qwen3-VL 4B & Causal & Yes & $-0.099$ & $-0.016$           & $-0.003$         & $+0.001$ \\
Qwen3-VL 8B & Causal & Yes & $-0.115$ & $-0.015$           & $+0.007$         & $-0.006$ \\
\midrule
PLM 1B      & Causal & No  & $-0.076$ & $\phantom{-}0.000$ & ---              & $\phantom{-}0.000$ \\
PLM 3B      & Causal & No  & $-0.115$ & $\phantom{-}0.000$ & ---              & $\phantom{-}0.000$ \\
\bottomrule
\end{tabular}}
\end{table}

\begin{table}[h]
\centering
\small
\caption{Activation patching across architectures. Molmo2: bidirectional LM attention over visual tokens; Qwen3-VL: causal LM attention. Layer-0 patching fully recovers Molmo at all three scales (act gap = 0.000, omitted); Qwen residual closes at layers~0--2 (DeepStack injection depth). Early/late boundary: L0--15/L16--31 for Molmo 4B/8B; L0--17/L18--35 for Molmo O-7B and Qwen. Baselines are this suite's own seed-42 runs and may differ from Appendix~\ref{app:tempcompass} tables by up to $0.007$.}
\label{tab:patching}
\resizebox{\columnwidth}{!}{%
\begin{tabular}{lcccccccc}
\toprule
\textbf{Condition} & \multicolumn{3}{c}{\textbf{Molmo2 (Acc)}} & \multicolumn{2}{c}{\textbf{Qwen3-VL 4B}} & \multicolumn{2}{c}{\textbf{Qwen3-VL 8B}} \\
\cmidrule(lr){2-4}\cmidrule(lr){5-6}\cmidrule(lr){7-8}
& 4B & 8B & O-7B & Acc & Gap & Acc & Gap \\
\midrule
\texttt{base}        & 0.728 & 0.730 & 0.718 & 0.703 & ---              & 0.738 & ---      \\
\texttt{shuf}        & 0.545 & 0.545 & 0.547 & 0.568 & $-0.135$         & 0.570 & $-0.168$ \\
\texttt{shuf-corr}   & 0.727 & 0.723 & 0.668 & 0.612 & $-0.091$         & 0.619 & $-0.119$ \\
\texttt{act\_layer0} & 0.728 & 0.730 & 0.718 & 0.687 & $-0.016$         & 0.723 & $-0.015$ \\
\texttt{act\_ds}     & ---   & ---   & ---   & 0.706 & $-0.003$         & 0.736 & $+0.007$ \\
\texttt{act\_early}  & 0.728 & 0.730 & 0.718 & 0.704 & $+0.001$         & 0.732 & $-0.006$ \\
\texttt{act\_late}   & 0.693 & 0.699 & 0.570 & 0.655 & $-0.048$         & 0.695 & $-0.043$ \\
\texttt{act\_all}    & 0.728 & 0.730 & 0.718 & 0.703 & $\phantom{-}0.000$ & 0.732 & $-0.006$ \\
\bottomrule
\end{tabular}
}
\par\vspace{4pt}
\end{table}

% ------------------------------------------------------------
\section{Conv3d Patch Embedding: Pre- vs.\ Post-Embedding Shuffle}
\label{app:patch-stage}
% ------------------------------------------------------------

Qwen3-VL's vision encoder uses a 3D Conv patch embedding (kernel $[2, 16, 16]$) that pairs two consecutive raw frames into one temporal unit, potentially encoding local motion prior to any attention. This raises the question of whether shuffling frame order \emph{before} the Conv3d embedding (as in our main conditions \texttt{shuf}/\texttt{shuf-corr}) captures the same temporal disruption as shuffling \emph{after} embedding (\texttt{emb-shuf}, embed-then-shuffle).

\paragraph{Procedure.}
Condition \texttt{emb-shuf} runs a \texttt{base} forward pass, captures the final ViT output embeddings and DeepStack tensors via forward hooks, then shuffles those embeddings by the same permutation used in \texttt{shuf} and injects them into a second generate pass that bypasses the ViT entirely. This isolates the effect of frame-order disruption at the LLM level without recomputing the ViT; any difference between \texttt{emb-shuf} and \texttt{shuf} (pixel-level shuffle, pre-embedding) would indicate that Conv3d motion signal contributes independently. Two recovery variants build on \texttt{emb-shuf}: \texttt{emb-shuf-ts} additionally remaps text timestamps to match the shuffle permutation; \texttt{emb-shuf-corr} additionally corrects temporal RoPE positions.

\paragraph{Results.}

\begin{table*}[t]
\centering
\small
\caption{Pre- vs.\ post-embedding shuffle on TempCompass (410 videos, single question per temporal aspect). Recovery conditions (\texttt{emb-shuf-ts}, \texttt{emb-shuf-corr}) test channel recovery on top of shuffled embeddings.}
\label{tab:embed-shuffle}
\begin{tabular}{llcc}
\toprule
\textbf{Condition} & \textbf{Description} & \textbf{4B} & \textbf{8B} \\
\midrule
\texttt{base}          & correct order, baseline             & 0.711 & 0.749 \\
\texttt{shuf}          & pixel shuffle before ViT            & 0.553 & 0.561 \\
\texttt{emb-shuf}      & shuffle ViT embeddings after ViT    & 0.561 & 0.567 \\
\texttt{emb-shuf-ts}   & \texttt{emb-shuf} + timestamp correction   & 0.621 & 0.653 \\
\texttt{emb-shuf-corr} & \texttt{emb-shuf} + RoPE correction        & 0.587 & 0.579 \\
\midrule
\texttt{emb-shuf} $-$ \texttt{shuf}              & & $+0.008$ & $+0.006$ \\
\texttt{emb-shuf-ts} $-$ \texttt{emb-shuf}       & & $+0.060$ & $+0.086$ \\
\texttt{emb-shuf-corr} $-$ \texttt{emb-shuf}     & & $+0.026$ & $+0.012$ \\
\bottomrule
\end{tabular}
\end{table*}

The gap between \texttt{emb-shuf} and \texttt{shuf} is $+0.006$--$+0.008$, within the range of run-to-run variability. We treat \texttt{emb-shuf} $\approx$ \texttt{shuf} as confirmation that our pixel-level shuffle conditions (\texttt{shuf}/\texttt{shuf-corr}) are equivalent to post-embedding shuffle for the purpose of temporal disruption, and that the temporal motion captured by Conv3d does not constitute an independent recoverable signal after shuffling. Within this experiment, correcting text timestamps on shuffled embeddings (\texttt{emb-shuf-ts}) recovers $+0.060$/$+0.086$ across model sizes, larger than RoPE correction alone (\texttt{emb-shuf-corr}: $+0.026$/$+0.012$), consistent with the timestamp-channel dominance reported in Section~\ref{sec:behavior:subset}.

The residual between \texttt{act\_layer0} patching and full baseline recovery in Qwen3-VL ($-0.016/-0.015$, $\approx$1~pp) is caused by DeepStack: the pre-hook fires at the \emph{input} of LM layer~0, but DeepStack adds shuffled intermediate ViT features \emph{inside} each of layers~0--2's forward passes, re-contaminating after each hook. A targeted \texttt{act\_ds} condition (patch layers~\{0,1,2\} only, exactly the DeepStack injection depth) recovers the full residual at both Qwen scales (\texttt{act\_ds} $-$ \texttt{act\_layer0} $= +0.016/+0.014$ at 4B/8B; \texttt{act\_early} $-$ \texttt{act\_ds} $\approx 0.000$ at both; $n=1580$, seed~42), confirming that layers~0--2 are the complete and sole source and that patching beyond layer~2 adds nothing. The present experiment confirms that Conv3d is not the source of this residual (\texttt{emb-shuf} $\approx$ \texttt{shuf}), isolating DeepStack as the explanation.

A full-scale replication (4B, $n=1580$) confirms \texttt{emb-shuf} $\approx$ \texttt{shuf} ($-0.003$, consistent with the prior $+0.008$ within run-to-run variation) and extends the experiment with condition \texttt{act\_layer0\_e1}: layer-0 activation patching applied on top of \texttt{emb-shuf} inputs. The gap \texttt{act\_layer0\_e1} $-$ \texttt{act\_layer0} $= -0.003$ ($0.691$ vs.\ $0.694$). patching layer~0 with Q0 hidden states gives the same recovery regardless of whether the ViT processed shuffled pixels (Q3) or correct pixels (E1). This directly isolates the residual to sequence ordering and DeepStack injection, independent of ViT pixel-encoding quality.

% ------------------------------------------------------------
\section{MRoPE Axis-Specific RoPE Corrections}
\label{app:axis-decomposition}
% ------------------------------------------------------------

Qwen3-VL uses 3D MRoPE with three axes: temporal (T), height (H), and width (W). All three axes share a per-frame base $t_i$; spatial offsets $h$ and $w$ are added within the frame. Under shuffle, the token at slot $j$ (containing frame $\pi(j)$) receives slot $j$'s temporal base $t_j$ across all three axes. \texttt{shuf-corr} reassigns the temporal base to $t_{\pi(j)}$ across all axes. \texttt{shuf-corr-T} and \texttt{shuf-corr-HW} correct only the T axis or only the H/W axes respectively, allowing the contribution of each axis to be read from the accuracy increment.

\paragraph{Results.}

\begin{table}[h]
\centering
\small
\caption{Axis-specific MRoPE correction on TempCompass ($n=1580$). Recovery scales with the number of corrected axes, consistent with the temporal base $t_i$ being encoded redundantly across all three MRoPE axes. \texttt{base-Trem} removes the temporal base from H/W axes under correct pixel order (sanity check; near \texttt{base}).}
\label{tab:axis-decomp}
\resizebox{\columnwidth}{!}{%
\begin{tabular}{lcccc}
\toprule
\textbf{Condition} & \textbf{Axes corrected} & \textbf{4B} & \textbf{8B} & \textbf{$\Delta$ \texttt{shuf} (4B/8B)} \\
\midrule
\texttt{shuf}           & 0 & 0.558 & 0.561 & --- \\
\texttt{shuf-corr-T}    & 1 & 0.570 & 0.575 & $+0.012$ / $+0.014$ \\
\texttt{shuf-corr-HW}   & 2 & 0.594 & 0.611 & $+0.036$ / $+0.050$ \\
\texttt{shuf-corr}      & 3 & 0.611 & 0.628 & $+0.053$ / $+0.067$ \\
\midrule
\texttt{base}           & --- & 0.710 & 0.743 & --- \\
\texttt{base-Trem}      & --- & 0.709 & 0.736 & $-0.001$ / $-0.007$ \\
\bottomrule
\end{tabular}
}
\end{table}

Recovery increments are monotone in the number of corrected axes ($+0.012 < +0.036 < +0.053$ for 4B). The H/W axes contribute more than the T axis alone ($+0.036$ vs $+0.012$), but neither reaches \texttt{shuf-corr} alone; all three together are needed for full (partial) recovery. This is consistent with M-RoPE's design: $t_i$ is the base for all three axes, so correcting any subset partially restores the per-frame temporal identity signal.

\paragraph{RoPE inter-frame spacing warp.}
Holding pixels and frame order fixed, we scale the inter-frame RoPE gap by a factor $\Delta$ (identity at $\Delta{=}1$, reproduced token-for-token; timestamps stripped) to test whether the model reads playback rate from positional spacing. Order and attribute\_change degrade under large stretch while speed and direction are unmoved at every $\Delta$ (Table~\ref{tab:warp}, Qwen3-VL 4B; the pattern replicates on Qwen3-VL 8B and Molmo2 4B/8B). The largest-$\Delta$ degradation partly reflects positions pushed toward the edge of the trained RoPE range; a spacing-only control isolating gap size from absolute magnitude is left to future work.

\begin{table}[h]
\centering
\small
\caption{RoPE inter-frame spacing warp (Qwen3-VL 4B, timestamps stripped; identity at $\Delta{=}1$). Distorting the spacing magnitude degrades order-dependent aspects, most under stretch; speed and direction are unmoved.}
\label{tab:warp}
\begin{tabular}{lcccc}
\toprule
$\Delta$ & speed & direction & order & attr\_change \\
\midrule
0.25 & 0.464 & 0.513 & 0.728 & 0.760 \\
0.5  & 0.486 & 0.519 & 0.738 & 0.778 \\
1    & 0.492 & 0.493 & 0.742 & 0.771 \\
2    & 0.508 & 0.510 & 0.732 & 0.767 \\
4    & 0.495 & 0.490 & 0.715 & 0.722 \\
8    & 0.479 & 0.496 & 0.675 & 0.677 \\
\bottomrule
\end{tabular}
\end{table}

% ------------------------------------------------------------
\section{DeepStack Ablation}
\label{app:deepstack}
% ------------------------------------------------------------

Qwen3-VL injects intermediate ViT features (from layers 8, 16, 24) into LLM layers 0--2 via residual addition at visual token positions (DeepStack). To test whether DeepStack carries an independent temporal signal not present in the main ViT output pathway, we ablate it (D-conditions: DeepStack replaced with identity) and correct it (Qc-conditions: inverse permutation applied to re-align shuffled intermediate features).

\begin{table}[h]
\centering
\small
\caption{DeepStack ablation on TempCompass ($n=1580$). \texttt{base-nds}/\texttt{shuf-nds}/\texttt{shuf-corr-nds} = DeepStack off; \texttt{base}/\texttt{shuf}/\texttt{shuf-corr} = DeepStack on. \texttt{shuf-dsc}/\texttt{shuf-corr-ds} = DeepStack corrected (inverse-permutation routing) with pixels shuffled. DeepStack adds $\approx$1 point at baseline and does not explain the shuffle residual gap.}
\label{tab:deepstack}
\resizebox{\columnwidth}{!}{%
\begin{tabular}{lcccc}
\toprule
\textbf{Condition} & \textbf{4B} & \textbf{8B} & \textbf{$\Delta$ vs DS-on (4B)} & \textbf{$\Delta$ vs DS-on (8B)} \\
\midrule
\texttt{base}           & 0.710 & 0.743 & --- & --- \\
\texttt{base-nds}       & 0.699 & 0.734 & $-0.011$ & $-0.009$ \\
\midrule
\texttt{shuf}           & 0.558 & 0.561 & --- & --- \\
\texttt{shuf-nds}       & 0.549 & 0.572 & $-0.009$ & $+0.011$ \\
\texttt{shuf-dsc}       & 0.572 & 0.587 & $+0.014$ & $+0.026$ \\
\midrule
\texttt{shuf-corr}      & 0.611 & 0.628 & --- & --- \\
\texttt{shuf-corr-nds}  & 0.602 & 0.649 & $-0.009$ & $+0.021$ \\
\texttt{shuf-corr-ds}   & 0.610 & 0.635 & $-0.001$ & $+0.007$ \\
\bottomrule
\end{tabular}
}
\end{table}

\paragraph{Interpretation.}
DeepStack contributes a small positive effect at baseline on TempCompass overall ($-0.011/-0.009$ when ablated) and a similarly small effect under shuffle; on TVBench \emph{moving\_direction} the ablation costs $-0.14$/$-0.06$ (Appendix~\ref{app:tvbench}), matching the larger layer-0 patching residual there. Correcting DeepStack routing (\texttt{shuf-dsc} vs.\ \texttt{shuf}: $+0.014/+0.026$) does not close the shuffle gap. The input-space residual gap remains large without DeepStack: \texttt{shuf-corr}$-$\texttt{base} is $-0.099/-0.115$ with DeepStack, whereas \texttt{shuf-corr-nds}$-$\texttt{base-nds} is $-0.097/-0.085$ without it. Thus, removing DeepStack changes the residual by only $+0.002$ at 4B and $+0.030$ at 8B and does not eliminate it. This indicates that DeepStack is not the independent temporal information channel responsible for the input-space \texttt{shuf-corr} gap, the result is instead consistent with the gap arising primarily from causal processing in the LLM.

\paragraph{Two residuals are not the same.}
This experiment concerns the \emph{input-space} \texttt{shuf-corr} residual ($\approx 10$ pp) and shows DeepStack does not explain it. Separately, in Section~\ref{sec:behavior:patching}, the much smaller $\approx 1.5$ pp \texttt{act\_layer0} \emph{patching} residual on Qwen3-VL is closed by \texttt{act\_ds}: during activation patching, DeepStack reinjects shuffled intermediate ViT features at vision-token positions \emph{after} the layer-0 input pre-hook fires, contaminating the patched run at layers~0--2. The two findings are compatible: DeepStack is not an independent temporal signal in normal forward operation, and is also the source of the small patching residual specifically because of when its injections fire relative to our pre-hook. Full per-condition TVBench numbers for DeepStack-off conditions are included in Appendix~\ref{app:tvbench}. In recovery terms, input-space RoPE correction (\texttt{shuf-corr}) closes only $29$--$33\%$ of the shuffle-to-baseline gap in Qwen, whereas activation patching closes $96$--$100\%$. The remainder is causally-mixed content that positional correction alone cannot reorganize. In Molmo both reach the same ceiling, consistent with bidirectional attention leaving no content-side contamination.

% ------------------------------------------------------------
\section{RoPE Depth Separation: Content vs.\ Positional Channels}
\label{app:rope-depth}
% ------------------------------------------------------------

The activation-patching experiments in Appendix~\ref{app:patching-full} establish that visual content order (via DeepStack) commits at LM layers~0--2. We ask where the RoPE positional effect accumulates by constructing a coherent conflict: correct visual content with fully reversed MRoPE positions, then patching early layers to see whether the RoPE damage is upstream or downstream of the DeepStack window.

\paragraph{Conditions.}
\begin{itemize}[leftmargin=*,topsep=2pt,itemsep=1pt]
  \item \texttt{rop-rev}: correct pixels, all three MRoPE axes deterministically reversed ($[T{-}1,\ldots,0]$ temporal, reversed H/W). Since pixels are correct, ViT outputs and DeepStack intermediate features are identical to \texttt{base}; the conflict is purely between correct visual content and reversed positional encoding.
  \item \texttt{act-ds-rop-rev}: \texttt{rop-rev} base with pre-hook input patching at LM layers~0--2. Because \texttt{rop-rev} and \texttt{base} share identical pixels their layer-0 inputs are identical; the layer-0 pre-hook is a no-op. Effective correction is layers~1--2 only.
  \item \texttt{act-out-ds-rop-rev}: \texttt{rop-rev} base with output patching at LM layers~0--2, implemented at the input of layer $i{+}1$ so it fires after the model loop's DeepStack additions. Installs the exact Q0 hidden state entering layer~3, verified bit-exact at the visual slots; reversed RoPE runs only at layers~3--35.
\end{itemize}

\begin{table}[h]
\centering
\small
\caption{RoPE depth separation on TempCompass ($n=1580$, seed~42, Qwen3-VL). \texttt{act\_ds} (from Appendix~\ref{app:patching-full}) included as reference: same patch depth over a \texttt{shuf} base gives full recovery. \texttt{act-ds-rop-rev} and \texttt{act-out-ds-rop-rev} apply input- and output-patching respectively over a \texttt{rop-rev} base.}
\label{tab:rope-depth}
\resizebox{\columnwidth}{!}{%
\begin{tabular}{lcccc}
\toprule
\textbf{Condition} & \textbf{4B} & \textbf{$\Delta$ base (4B)} & \textbf{8B} & \textbf{$\Delta$ base (8B)} \\
\midrule
\texttt{base}                  & 0.703 & ---      & 0.737 & ---      \\
\texttt{rop-rev}               & 0.655 & $-0.048$ & 0.675 & $-0.063$ \\
\texttt{act\_ds}               & 0.704 & $+0.001$ & 0.737 & $-0.001$ \\
\texttt{act-ds-rop-rev}        & 0.645 & $-0.058$ & 0.678 & $-0.059$ \\
\texttt{act-out-ds-rop-rev}    & 0.655 & $-0.048$ & 0.680 & $-0.057$ \\
\bottomrule
\end{tabular}
}
\par\vspace{4pt}
{\footnotesize
\textit{Key comparisons.}
\texttt{act-ds-rop-rev} $-$ \texttt{rop-rev} $\approx 0.000$ at both sizes: input-patching layers~0--2 provides no benefit against reversed RoPE active inside those layers.
\texttt{act-out-ds-rop-rev} $-$ \texttt{rop-rev} $= +0.000$/$+0.006$: output-patching, installing the bit-exact Q0 state entering layer~3, provides no benefit either.
\texttt{act-out-ds-rop-rev} $-$ \texttt{act\_ds} $= -0.049$/$-0.056$ at 4B/8B: reversed RoPE at layers~3--35 alone, from a provably clean Q0 residual at layer~3, reproduces the full \texttt{rop-rev} accuracy cost.}
\end{table}

Both input and output-patching are null results, localizing the RoPE accuracy damage entirely to layers~3--35. The contrast with \texttt{act\_ds} (same patch depth, \texttt{shuf} base, full recovery) confirms the asymmetry: content-ordering errors are remediable at layers~0--2, positional-encoding errors are not. The probe-decay analysis of Appendix~\ref{app:probe-detail} explains why: the linearly decodable frame-order code is not present in the early residual to be protected, it is constructed across the middle layers by position-driven attention, so a clean state entering layer~3 cannot rescue a construction that runs with reversed coordinates.

% ------------------------------------------------------------
\section{Timestamp-Sensitivity Partition: Per-Aspect Breakdowns}
\label{app:subset-aspect}
% ------------------------------------------------------------

Section~\ref{sec:behavior:subset} summarizes the timestamp-sensitivity partition. Table~\ref{tab:subset-conflict} gives the pooled results, and Table~\ref{tab:subset-aspect} the per-aspect breakdowns for aspects where $|S_\mathrm{ts}| \geq 20$. Aspects below this threshold are suppressed; bootstrap CIs use 1000 resamples.

\begin{table}[h]
\centering
\small
\caption{Timestamp-sensitivity partition: subset sizes ($n$) and \texttt{conf-ts} accuracy. $S_\mathrm{ts}$: correct under \texttt{base}, incorrect under \texttt{base-nts}. $S_\mathrm{neither}$: correct under both. 95\% bootstrap CIs in Table~\ref{tab:subset-aspect}.}
\label{tab:subset-conflict}
\begin{tabular}{lccccc}
\toprule
\textbf{Model} & \multicolumn{2}{c}{\boldmath$S_\mathrm{ts}$} & \multicolumn{2}{c}{\boldmath$S_\mathrm{neither}$} & \boldmath$\Delta$ \\
\cmidrule(lr){2-3}\cmidrule(lr){4-5}
& $n$ & Acc & $n$ & Acc & \\
\midrule
Molmo2 4B   & 93  & 0.677 & 1058 & 0.987 & $+0.309$ \\
Molmo2 8B   & 168 & 0.815 & 985  & 0.986 & $+0.170$ \\
Qwen3-VL 4B & 78  & 0.487 & 1044 & 0.936 & $+0.449$ \\
Qwen3-VL 8B & 98  & 0.704 & 1063 & 0.977 & $+0.273$ \\
\bottomrule
\end{tabular}
\end{table}

\begin{table*}[t]
\centering
\small
\caption{Per-aspect conflict-condition (\texttt{conf-ts}: forward pixels, reversed timestamps) accuracy on $S_\mathrm{ts}$ and $S_\mathrm{neither}$ subsets. Aspects shown only where $|S_\mathrm{ts}| \geq 20$. Numbers below $n=20$ are suppressed (---).}
\label{tab:subset-aspect}
\begin{tabular}{llcccc}
\toprule
\textbf{Model} & \textbf{Aspect} & \textbf{Acc}$|S_\mathrm{ts}$ & $n$ & \textbf{Acc}$|S_\mathrm{nei}$ & $n$ \\
\midrule
\multirow{4}{*}{Molmo2 4B}
  & attribute\_change & 0.591 [0.409, 0.818] & 22 & 0.995 [0.986, 1.000] & 214 \\
  & direction         & 0.650 [0.450, 0.850] & 20 & 0.970 [0.939, 0.994] & 165 \\
  & speed             & 0.771 [0.629, 0.886] & 35 & 0.985 [0.963, 1.000] & 134 \\
  & \textbf{Pooled}   & \textbf{0.677} [0.581, 0.774] & \textbf{93} & \textbf{0.987} [0.979, 0.993] & \textbf{1058} \\
\midrule
\multirow{5}{*}{Molmo2 8B}
  & attribute\_change & 0.875 [0.771, 0.958] & 48 & 0.995 [0.984, 1.000] & 193 \\
  & direction         & 0.923 [0.821, 1.000] & 39 & 0.993 [0.978, 1.000] & 138 \\
  & order             & 0.531 [0.344, 0.719] & 32 & 0.965 [0.941, 0.990] & 202 \\
  & speed             & 0.864 [0.750, 0.955] & 44 & 0.968 [0.937, 1.000] & 126 \\
  & \textbf{Pooled}   & \textbf{0.815} [0.750, 0.869] & \textbf{168} & \textbf{0.986} [0.979, 0.992] & \textbf{985} \\
\midrule
\multirow{3}{*}{Qwen3-VL 4B}
  & direction         & 0.478 [0.304, 0.696] & 23 & 0.952 [0.911, 0.979] & 146 \\
  & speed             & 0.600 [0.433, 0.767] & 30 & 0.935 [0.891, 0.971] & 138 \\
  & \textbf{Pooled}   & \textbf{0.487} [0.372, 0.603] & \textbf{78} & \textbf{0.936} [0.921, 0.950] & \textbf{1044} \\
\midrule
\multirow{3}{*}{Qwen3-VL 8B}
  & direction         & 0.560 [0.360, 0.720] & 25 & 0.938 [0.897, 0.972] & 145 \\
  & speed             & 0.805 [0.683, 0.927] & 41 & 0.958 [0.924, 0.986] & 144 \\
  & \textbf{Pooled}   & \textbf{0.704} [0.612, 0.796] & \textbf{98} & \textbf{0.977} [0.968, 0.986] & \textbf{1063} \\
\bottomrule
\end{tabular}
\end{table*}

Several patterns are worth noting beyond the pooled results. First, the action aspect contributes negligibly to $S_\mathrm{ts}$ across all four models (0--5 questions), because action accuracy is near ceiling regardless of timestamp presence; the aspect is not informative for this partition. Second, the order aspect enters $S_\mathrm{ts}$ only for Molmo2 8B ($n=32$, Acc$|S_\mathrm{ts}=0.531$), with a $\Delta$ of $+0.434$, among the largest single-aspect gaps in the table. Molmo2 8B specifically appears to use timestamps to judge event order, and reversed timestamps are strongly misleading on those questions. Third, the $S_\mathrm{ts}$ partitions are model-specific by construction: Qwen's timestamp-reliant questions are concentrated in direction and speed, while Molmo 4B's include attribute\_change. The four models do not agree on which questions require timestamps, consistent with timestamp-dependence being a property of how each model uses the channel.

% ------------------------------------------------------------
\section{Full Probe Results}
\noindent\textit{Supports Secs.~\ref{sec:behavior:patching} and~\ref{sec:discussion:channels}: probe methodology, the probe-decay (constructibility) analysis, and cross-condition transfer.}
\label{app:probe-detail}
% ------------------------------------------------------------

\paragraph{Probe method.} We train multi-class logistic regression probes on residual-stream activations mean-pooled over each frame's vision span, at LM layers $L_0, L_4, L_8, \ldots$, targeting two quantities that coincide under unshuffled inputs and dissociate under shuffle: the \emph{original-frame index} (frame target) and the \emph{slot index} (slot target). Cross-condition transfer trains on \texttt{shuf} or \texttt{shuf-corr} and evaluates on \texttt{base} to test whether a readout direction is condition-specific or general. Splits and bootstrap CIs follow the conventions in Section~\ref{sec:behavior:patching}. Probes are applied to Qwen3-VL and Molmo2 only. A complementary probe trained on only the frame-delimiter (first) token is reported in App.~\ref{app:first-token}.

\subsection*{Frame-Identity and Slot Readout Under Disruption}

The \texttt{base} frame-identity probe peaks at strong accuracy in both architectures: 0.83/0.81 in Qwen 4B/8B at L20/L35; 0.42/0.46/0.80 in Molmo 4B/8B/O-7B at L20/L20/L12 (Table~\ref{tab:probe-l20}). Frame identity is linearly recoverable from baseline mean-pooled activations through most of the network, peaking in middle-to-late layers.

Under \texttt{shuf}, frame identity collapses to chance in both architectures: $\le 0.02$ across all sizes. Under \texttt{shuf-corr}, frame identity is partially recovered, substantially more in Molmo than Qwen: gap closed is 74\%/64\%/30\% (Molmo 4B/8B/O-7B) vs.\ 17\%/20\% (Qwen 4B/8B). Bidirectional LM attention permits more representational recovery than causal LM attention does, paralleling the behavioral recovery pattern of Section~\ref{sec:behavior:patching}.

Under \texttt{shuf}, the slot-identity probe trained on \texttt{base} activations remains at near-baseline accuracy in both architectures (Qwen $0.800/0.785$; Molmo $0.402/0.451/0.783$). Under \texttt{shuf-corr}, the slot readout collapses to chance in Molmo ($0.040/0.042/0.045$) and near-chance in Qwen ($0.216/0.204$). Correcting RoPE to follow original frame indices destroys the slot-organized readout direction; Molmo's tighter RoPE anchoring makes this collapse more complete than in Qwen, where causal sequential ordering provides residual slot signal.

\paragraph{RoPE disruption with correct pixels damages the readout.}
The \texttt{rop-shuf} and \texttt{rop-rev} rows of Table~\ref{tab:probe-l20} isolate the contribution of RoPE to the readout: visual content is intact but positional encoding is shuffled or reversed. Frame-probe accuracy drops substantially in Qwen (\texttt{base}$-$\texttt{rop-shuf} $= 0.629/0.610$ at 4B/8B; \texttt{base}$-$\texttt{rop-rev} $= 0.651/0.617$) and to chance in Molmo (\texttt{rop-shuf} frame-probe $0.044/0.047/0.048$ across 4B/8B/O-7B). The \texttt{base}-trained probe reads off a representation that depends on RoPE matching the position the model expects; when RoPE is wrong, the readout direction does not recover frame identity even with correct visual content. This is the representational counterpart to the behavioral \texttt{rop-shuf} effect in Section~\ref{sec:behavior:rope}.

\paragraph{Per-layer trajectories.}
Table~\ref{tab:probe-layers} gives the layer-by-layer frame-identity probe for
all five models under \texttt{base} and \texttt{rop-shuf} (correct pixels,
shuffled positions), from the same runs as Table~\ref{tab:probe-l20}. Every
model starts at or near chance at the LM input and builds the order code
across the middle layers; under shuffled positions the code is never built in
Molmo at any scale, while Qwen retains a weak above-chance residue.

\begin{table}[h]
\centering
\small
\caption{Frame-identity probe accuracy by layer (mean-pooled, trained on
\texttt{base} at each layer). \emph{peak}: the best \texttt{base} layer
(L20 for Qwen 4B and Molmo 4B/8B, L35 for Qwen 8B, L12 for O-7B); the
\texttt{rop-shuf} peak column reports that same layer. Chance is $0.090$/$0.091$
for Qwen 4B/8B and $0.041$/$0.042$/$0.041$ for Molmo 4B/8B/O-7B.}
\label{tab:probe-layers}
\resizebox{\columnwidth}{!}{%
\begin{tabular}{llccccccc}
\toprule
\textbf{Model} & \textbf{Cond.} & L0 & L4 & L8 & L12 & L16 & L20 & peak \\
\midrule
Qwen 4B  & \texttt{base}     & 0.121 & 0.438 & 0.600 & 0.619 & 0.814 & 0.830 & 0.830 \\
         & \texttt{rop-shuf} & 0.085 & 0.217 & 0.135 & 0.125 & 0.184 & 0.201 & 0.201 \\
Qwen 8B  & \texttt{base}     & 0.277 & 0.547 & 0.654 & 0.669 & 0.778 & 0.755 & 0.810 \\
         & \texttt{rop-shuf} & 0.110 & 0.178 & 0.149 & 0.157 & 0.218 & 0.220 & 0.200 \\
Molmo 4B & \texttt{base}     & 0.059 & 0.105 & 0.158 & 0.278 & 0.332 & 0.416 & 0.416 \\
         & \texttt{rop-shuf} & 0.057 & 0.048 & 0.050 & 0.045 & 0.039 & 0.044 & 0.044 \\
Molmo 8B & \texttt{base}     & 0.052 & 0.065 & 0.127 & 0.122 & 0.267 & 0.458 & 0.458 \\
         & \texttt{rop-shuf} & 0.051 & 0.044 & 0.048 & 0.046 & 0.040 & 0.047 & 0.047 \\
Molmo O-7B & \texttt{base}   & 0.151 & 0.234 & 0.694 & 0.799 & 0.762 & 0.679 & 0.799 \\
         & \texttt{rop-shuf} & 0.112 & 0.094 & 0.051 & 0.048 & 0.056 & 0.067 & 0.048 \\
\bottomrule
\end{tabular}}
\end{table}

\begin{table}[h]
\centering
\small
\caption{Probe accuracy at the best baseline-condition layer for each model, with 95\% bootstrap confidence intervals. Frame target = original-frame index; slot target = position-in-input. Best layer: L20 for Qwen 4B and Molmo 4B/8B; L35 for Qwen 8B; L12 for Molmo O-7B.}
\label{tab:probe-l20}
\resizebox{\columnwidth}{!}{%
\begin{tabular}{lcccc}
\toprule
\textbf{Model} & \multicolumn{2}{c}{\textbf{Frame target}} & \multicolumn{2}{c}{\textbf{Slot target}} \\
\textbf{/ Condition} & Acc & CI & Acc & CI \\
\midrule
\multicolumn{5}{l}{\emph{Qwen3-VL 4B (L20)}} \\
\texttt{base}      & 0.830 & [0.806, 0.854] & 0.830 & (same) \\
\texttt{shuf}      & 0.013 & [0.007, 0.021] & 0.800 & [0.774, 0.825] \\
\texttt{shuf-corr} & 0.139 & [0.117, 0.162] & 0.216 & [0.189, 0.243] \\
\texttt{rop-shuf}  & 0.201 & [0.175, 0.228] & 0.201 & (slot=frame) \\
\texttt{rop-rev}   & 0.179 & [0.155, 0.205] & 0.179 & (slot=frame) \\
\midrule
\multicolumn{5}{l}{\emph{Qwen3-VL 8B (L35)}} \\
\texttt{base}      & 0.810 & [0.785, 0.836] & 0.810 & (same) \\
\texttt{shuf}      & 0.010 & [0.004, 0.017] & 0.785 & [0.758, 0.811] \\
\texttt{shuf-corr} & 0.166 & [0.143, 0.191] & 0.204 & [0.177, 0.231] \\
\texttt{rop-shuf}  & 0.200 & [0.174, 0.227] & 0.200 & (slot=frame) \\
\texttt{rop-rev}   & 0.193 & [0.167, 0.220] & 0.193 & (slot=frame) \\
\midrule
\multicolumn{5}{l}{\emph{Molmo2 4B (L20)}} \\
\texttt{base}      & 0.416 & [0.394, 0.438] & 0.416 & (same) \\
\texttt{shuf}      & 0.023 & [0.016, 0.030] & 0.402 & [0.380, 0.423] \\
\texttt{rop-shuf}  & 0.044 & [0.035, 0.053] & 0.044 & ($\approx$ chance) \\
\texttt{shuf-corr} & 0.308 & [0.288, 0.329] & 0.040 & [0.032, 0.049] \\
\midrule
\multicolumn{5}{l}{\emph{Molmo2 8B (L20)}} \\
\texttt{base}      & 0.458 & [0.435, 0.481] & 0.458 & (same) \\
\texttt{shuf}      & 0.024 & [0.017, 0.031] & 0.451 & [0.429, 0.474] \\
\texttt{rop-shuf}  & 0.047 & [0.037, 0.056] & 0.047 & ($\approx$ chance) \\
\texttt{shuf-corr} & 0.292 & [0.271, 0.312] & 0.042 & [0.033, 0.052] \\
\midrule
\multicolumn{5}{l}{\emph{Molmo2 O-7B (L12)}} \\
\texttt{base}      & 0.799 & [0.781, 0.816] & 0.799 & (same) \\
\texttt{shuf}      & 0.008 & [0.004, 0.012] & 0.783 & [0.765, 0.801] \\
\texttt{rop-shuf}  & 0.048 & [0.039, 0.058] & 0.048 & ($\approx$ chance) \\
\texttt{shuf-corr} & 0.238 & [0.219, 0.257] & 0.045 & [0.036, 0.055] \\
\bottomrule
\end{tabular}
}
\end{table}

\subsection*{Probe Decay: the Order Code Is Constructed, Not Stored}

A frame-order probe (which original frame occupies slot $j$) trained on clean-run (\texttt{base}) layer-\emph{input} activations, 100 videos, 80/20 split, chance $0.098$, evaluated at every layer input under three conditions: \texttt{base}, \texttt{rop-rev} (correct pixels, reversed MRoPE), and \texttt{act-out-ds-rop-rev} (bit-exact clean residual installed entering layer~3, reversed RoPE beyond; Appendix~\ref{app:rope-depth}).

\begin{table}[h]
\centering
\small
\caption{Frame-order probe accuracy by layer input (Qwen3-VL-4B, mean-pooled visual span). The patched run matches \texttt{base} exactly through layer~3 and converges to the reversed-RoPE floor by layer~8.}
\label{tab:probe-decay}
\begin{tabular}{lccc}
\toprule
\textbf{Layer input} & \texttt{base} & \texttt{rop-rev} & \texttt{act-out} \\
\midrule
0  & 0.160 & 0.160 & 0.160 \\
3  & 0.272 & 0.218 & 0.272 \\
4  & 0.350 & 0.204 & 0.248 \\
6  & 0.408 & 0.184 & 0.218 \\
8  & 0.471 & 0.150 & 0.155 \\
20 & 0.835 & 0.180 & 0.175 \\
\bottomrule
\end{tabular}
\end{table}

Two readings. First, collapse is fast: the patched run tracks \texttt{base} exactly at layer~3 (inputs are bit-identical), falls behind at the first reversed-RoPE layer, and reaches the reversed-RoPE floor within about four layers, consistent with the behavioral null of Appendix~\ref{app:rope-depth}. Second, and more important: even in the clean run, order decodability is only $0.27$ entering layer~3 and is \emph{built} to $0.84$ by layer~20. The linearly decodable order code is not present early and destroyed by reversed RoPE; it is constructed across the middle layers by position-driven attention, and under reversed positions it is never built. Linear-probe accuracy measures decodability, not information content.

\subsection*{Cross-Condition Probe Transfer}

\begin{table}[h]
\centering
\small
\caption{Cross-condition probe transfer. The slot probe direction is general (\texttt{shuf}-trained probes transfer cleanly to \texttt{base}). The frame probe direction shows partial overlap between \texttt{shuf-corr} and \texttt{base}.}
\label{tab:probe-transfer}
\resizebox{\columnwidth}{!}{%
\begin{tabular}{llccc}
\toprule
\textbf{Probe} & \textbf{Source} & \textbf{Self acc} & \textbf{Eval on \texttt{base}} & \textbf{Retention} \\
\midrule
Slot, Qwen 4B    & \texttt{shuf} (L36) & 0.812 & 0.809 & 100\% \\
Slot, Qwen 8B    & \texttt{shuf} (L35) & 0.788 & 0.796 & 100\% \\
Slot, Molmo 4B   & \texttt{shuf} (L24) & 0.455 & 0.388 & 85\% \\
Slot, Molmo 8B   & \texttt{shuf} (L20) & 0.537 & 0.463 & 86\% \\
Slot, Molmo O-7B & \texttt{shuf} (L12) & 0.805 & 0.752 & 93\% \\
\midrule
Frame, Qwen 4B    & \texttt{shuf-corr} (L8)  & 0.511 & 0.466 & 91\% \\
Frame, Qwen 8B    & \texttt{shuf-corr} (L35) & 0.548 & 0.532 & 97\% \\
Frame, Molmo 4B   & \texttt{shuf-corr} (L28) & 0.307 & 0.307 & 100\% \\
Frame, Molmo 8B   & \texttt{shuf-corr} (L28) & 0.310 & 0.372 & 120\% \\
Frame, Molmo O-7B & \texttt{shuf-corr} (L12) & 0.578 & 0.407 & 70\% \\
\bottomrule
\end{tabular}
}
\end{table}

\begin{table}[h]
\centering
\small
\caption{Behavioral and representational recovery from shuffle via RoPE correction. Gap closed = (\texttt{shuf-corr}$-$\texttt{shuf})/(\texttt{base}$-$\texttt{shuf}). Rank order is identical across both measures.}
\label{tab:beh-vs-rep}
\resizebox{\columnwidth}{!}{%
\begin{tabular}{lcc}
\toprule
\textbf{Model} & \textbf{Behavioral gap closed} & \textbf{Frame-probe gap closed} \\
\midrule
Qwen3-VL 4B  & 35\% & 15\% \\
Qwen3-VL 8B  & 37\% & 20\% \\
Molmo2 4B    & 99\% & 73\% \\
Molmo2 8B    & 98\% & 62\% \\
Molmo2 O-7B  & 68\% & 29\% \\
\bottomrule
\end{tabular}
}
\end{table}

\subsection*{First-token delimiter probe}
\label{app:first-token}

The main probe analysis (Section~\ref{sec:behavior:patching}) uses mean-pooled activations across all tokens in each frame's vision span (frame delimiter plus visual patches). Here we report a complementary probe trained on the activation at only the first token of each frame's span, the frame delimiter token.

\paragraph{Motivation.}
Delimiter tokens sit at the boundary between frames in the input sequence and receive the frame's sequential position index via RoPE. Their activations therefore primarily reflect positional-embedding lookup rather than visual content aggregated across patches. Probing them serves as a sanity check: if our RoPE injection is working correctly, the delimiter probe should show near-perfect slot accuracy under native RoPE (the delimiter token's position encodes its slot) and near-perfect frame accuracy under corrected RoPE (the delimiter token's position encodes the original frame index). Failure of these checks would indicate the RoPE manipulation is not reaching the model's activations.

\paragraph{Results.}
At layer 0, the delimiter probe under \texttt{base} achieves near-perfect slot (and frame) accuracy ($\approx 0.988$), as expected: the first layer's delimiter activation is dominated by the positional embedding at that slot. Under \texttt{rop-shuf} (correct pixels, shuffled RoPE), delimiter accuracy at L0 drops to $\approx 0.011$ for the slot direction, confirming RoPE injection changes what the delimiter encodes. Under \texttt{shuf-corr} (shuffled pixels, corrected RoPE), delimiter accuracy at L0 is $0.749$ for the frame direction and $\approx 0.000$ for the slot direction, consistent with the corrected position encoding the original frame index rather than the input slot.

These checks confirm the RoPE injections are operating as intended at the input to the first LM layer. The delimiter probe is not used for the main probe analysis because delimiter activations are dominated by positional embedding lookup and do not reflect the distributed visual representation of the frame; mean pooled activations across the full vision span are a more informative measurement of how frame identity is encoded after visual processing.

% ------------------------------------------------------------
\section{Full Condition Reference}
\label{app:conditions-full}
% ------------------------------------------------------------

Table~\ref{tab:conditions-full} extends Table~\ref{tab:conditions-main} with all supplementary conditions used in appendix analyses and referenced in the main text but not listed in the main table.

\begin{table*}[t]
\centering
\small
\caption{Full condition reference. ``Native'' RoPE = positions match input order. ``Corrected'' = positions follow original frame indices. ``Shuffled/reversed'' = positions independently permuted or reversed regardless of pixel order.}
\label{tab:conditions-full}
\begin{tabular}{lllll}
\toprule
\textbf{Code} & \textbf{Pixels} & \textbf{Timestamps} & \textbf{RoPE} & \textbf{Notes} \\
\midrule
\multicolumn{5}{l}{\emph{Qwen3-VL (supplementary)}} \\
\texttt{shuf-corr-T}   & shuffled & at original slots & T axis corrected & MRoPE T axis only \\
\texttt{shuf-corr-HW}  & shuffled & at original slots & H/W axes corrected & MRoPE H/W only \\
\texttt{shuf-corr-ds}  & shuffled & at original slots & corrected (all) & + DeepStack corrected \\
\texttt{shuf-full}     & shuffled & moved with frames & corrected (all) & full correction \\
\texttt{base-Trem}     & original & original & T removed from H/W & detangle T from spatial axes \\
\texttt{base-nds}      & original & original & native & DeepStack off \\
\texttt{shuf-nds}      & shuffled & at original slots & native & DeepStack off \\
\texttt{shuf-corr-nds} & shuffled & at original slots & corrected & DeepStack off \\
\texttt{shuf-dsc}      & shuffled & at original slots & native & DeepStack routing corrected \\
\texttt{emb-shuf}      & embed-then-shuffle & at original slots & native & post-ViT temporal shuffle \\
\texttt{emb-shuf-ts}   & embed-then-shuffle & corrected to perm & native & emb-shuf + timestamp correction \\
\texttt{emb-shuf-corr} & embed-then-shuffle & at original slots & corrected & emb-shuf + RoPE correction \\
\texttt{rop-shuf}      & original & original & shuffled by perm & RoPE corrupt, pixels correct \\
\texttt{rop-rev}       & original & original & reversed & RoPE reversed, pixels correct \\
\texttt{rev}           & reversed & at original slots & native & reversal baseline \\
\texttt{rev-corr}      & reversed & at original slots & corrected & reversal + RoPE correct \\
\texttt{conf-ts}       & original & reversed & native & conflict: forward pix, rev ts \\
\texttt{rev-mvts}      & reversed & moved with frames & native & consistent reversal \\
\texttt{conf-ts2}      & shuffled (perm A) & shuffled (perm B) & native & conflict: independent perms \\
\midrule
\multicolumn{5}{l}{\emph{Molmo2 (supplementary)}} \\
\texttt{shuf-nts}  & shuffled & removed & native & shuffle, no timestamps \\
\texttt{rev}       & reversed & at original slots & native & reversal baseline \\
\texttt{rev-corr}  & reversed & at original slots & corrected & reversal + RoPE correct \\
\texttt{blank-nts} & blank & none & native & text floor (no visual, no ts) \\
\texttt{base-nts}  & original & absent & native & baseline without timestamps \\
\texttt{shuf-full} & shuffled & moved with frames & corrected & full correction \\
\texttt{conf-ts}   & original & reversed & native & conflict: forward pix, rev ts \\
\texttt{rev-mvts}  & reversed & moved with frames & native & consistent reversal \\
\texttt{conf-ts2}  & shuffled (perm A) & shuffled (perm B) & native & conflict: independent perms \\
\midrule
\multicolumn{5}{l}{\emph{PerceptionLM (supplementary)}} \\
\texttt{rev} & reversed & --- & native & reversal baseline \\
\bottomrule
\end{tabular}
\end{table*}

% ------------------------------------------------------------
\section{PerceptionLM Patching Results}
\label{app:plm}
% ------------------------------------------------------------

PerceptionLM (PLM) uses a LLaMA3 backbone with a per-frame independent ViT (no cross-frame attention, no DeepStack injections) and 1D RoPE. This makes it structurally distinct from both Molmo2 (bidirectional LM attention) and Qwen3-VL (causal LM + DeepStack), providing a third data point that separates the DeepStack and LM attention type mechanisms.

\begin{table}[h]
\centering
\small
\caption{PerceptionLM behavioral conditions (TempCompass, $n=1580$) and activation patching results. \texttt{rev} = reversed pixel order. \texttt{shuf-corr} = shuffle + RoPE corrected. Patching conditions: act\_layer0 = layer~0 only; act\_early = L0--7 (1B) / L0--13 (3B); act\_late = L8--15 (1B) / L14--27 (3B); act\_all = all layers. The \texttt{rev-corr} channel dissociation is reported separately in Table~\ref{tab:plm-dissoc}, since under \texttt{rev-corr} the paired \texttt{\_reverse} videos realize the mirror conflict.}
\label{tab:plm-full}
\begin{tabular}{lcc}
\toprule
\textbf{Condition} & \textbf{1B} & \textbf{3B} \\
\midrule
\texttt{base}         & 0.632 & 0.674 \\
\texttt{single-frame} & 0.459 & 0.466 \\
\texttt{shuf}         & 0.528 & 0.527 \\
\texttt{rev}          & 0.442 & 0.421 \\
\texttt{shuf-corr}    & 0.556 & 0.559 \\
\midrule
\texttt{act\_layer0}  & 0.632 & 0.674 \\
\texttt{act\_early}   & 0.632 & 0.674 \\
\texttt{act\_late}    & 0.577 & 0.531 \\
\texttt{act\_all}     & 0.632 & 0.674 \\
\midrule
\texttt{shuf-corr} $-$ \texttt{base} & $-0.076$ & $-0.115$ \\
\texttt{act\_layer0} $-$ \texttt{base} & $\phantom{-}0.000$ & $\phantom{-}0.000$ \\
\texttt{act\_late} $-$ \texttt{base}   & $-0.055$ & $-0.143$ \\
\bottomrule
\end{tabular}
\end{table}

\paragraph{Behavioral conditions.}
\texttt{shuf-corr} leaves a recovery gap of $-0.076/-0.115$ (1B/3B), comparable to Qwen's $-0.099/-0.115$, confirming that RoPE correction cannot undo the damage from causal LM processing of shuffled frame embeddings. This matches the pattern for all causal-LM models. The reversal damage ($-0.190/-0.253$) shows PLM does track temporal order (from positions in part, see below), despite having no timestamp channel in its prompts.

\paragraph{Channel dissociation (rev-corr).}
The reverse-plus-corrected-RoPE condition (\texttt{rev-corr}, reversed pixels with positions held forward) places PLM on the dominance axis of Section~\ref{sec:behavior:rope}, reported per scale in Table~\ref{tab:plm-dissoc} on \emph{attribute\_change} (the diagnostic paired aspect), original videos only. On the original videos, accuracy falls from \texttt{base} by $+0.167/+0.236$ (1B/3B). Scored against the forward ($E$) and reversed ($F$) ground truth on the $n=141$ informative pairs (original GT $\neq$ reversed GT), $F-E = -0.525/-0.489$ (motion-dependent slice $-0.545/-0.487$): under conflict PLM predominantly answers the forward event (at 1B a fraction $E=0.72$ answer forward and $F=0.19$ reversed, the rest a distractor; 3B is $0.71/0.22$), so it is a partial, position-leaning overrider. This is weaker than Molmo2 ($F-E$ near $-0.95$) and more position-leaning than Qwen3-VL ($-0.23$, and $+0.01/+0.24$ on the motion slice), placing PLM between the two and consistent with both positional encoding and attention type contributing (Section~\ref{sec:discussion:channels}). PLM 1B's forward answers under \texttt{rev-corr} cannot be semantic: on the reversed-clip control (Appendix~\ref{app:revclip}) its errors sit at the random baseline, so the forward pull under the conflict is positional. The plain \texttt{rev} row in Table~\ref{tab:plm-full} (reversed pixels, native RoPE) is a different condition and does not isolate the channel.

\begin{table}[h]
\centering
\small
\caption{PerceptionLM channel dissociation under \texttt{rev-corr} (reversed pixels, RoPE held forward), TempCompass \emph{attribute\_change}, original videos. \texttt{base}, \texttt{rev-corr}, and the drop are over all 144 paired items. $E$ (answers forward GT), $F$ (answers reversed GT), $F-E$, and the read breakdown (forward / reversed / distractor) are over the $n$ informative pairs (original GT $\neq$ reversed GT). $F-E<0$ marks order read from RoPE rather than pixels. Values are per scale, not pooled.}
\label{tab:plm-dissoc}
\resizebox{\columnwidth}{!}{%
\begin{tabular}{lccccccc}
\toprule
\textbf{Scale} & \texttt{base} & \texttt{rev-corr} & \textbf{drop} & $E$ & $F$ & $F\!-\!E$ & \textbf{fwd/rev/conf ($n$)} \\
\midrule
1B & 0.875 & 0.708 & $+0.167$ & 0.716 & 0.191 & $-0.525$ & 101/27/13 (141) \\
3B & 0.938 & 0.701 & $+0.236$ & 0.709 & 0.220 & $-0.489$ & 100/31/10 (141) \\
\bottomrule
\end{tabular}}
\end{table}

\paragraph{Activation patching.}
\texttt{act\_layer0} $=$ \texttt{base} exactly at both scales, in contrast to Qwen's $-0.015/-0.016$ residual. PLM has no DeepStack injections, so the layer-0 pre-hook captures exactly the point at which all visual features enter the LM; no subsequent injection can re-introduce shuffled order information. \texttt{act\_early} $=$ \texttt{act\_all} $=$ \texttt{base}, confirming no further gain from patching deeper layers once the correct per-frame features are in place.

\texttt{act\_late} is notably below baseline, especially at 3B ($-0.143$, nearly at the shuffled-input floor). With 14 early layers processing shuffled frame embeddings causally before the late-patch window begins, early-layer contamination is largely committed and cannot be undone by late patching. The depth-dependence of this penalty (1B: $-0.055$; 3B: $-0.143$) confirms that deeper causal LMs accumulate order information more pervasively.

% ------------------------------------------------------------
\section{Full TempCompass Results}
\label{app:tempcompass}
% ------------------------------------------------------------

Tables~\ref{tab:tempcompass-molmo}--\ref{tab:tempcompass-qwen8b} report TempCompass accuracy for Molmo2 and Qwen3-VL across all behavioral conditions referenced in the main text. PerceptionLM behavioral results are in Table~\ref{tab:plm-full} (Appendix~\ref{app:plm}). MRoPE axis-specific corrections (\texttt{shuf-corr-T}, \texttt{shuf-corr-HW}, \texttt{shuf-corr-ds}) and DeepStack-off conditions (\texttt{base-nds}, \texttt{shuf-nds}, \texttt{shuf-corr-nds}) for Qwen3-VL are tabulated separately in Appendices~\ref{app:axis-decomposition} and~\ref{app:deepstack} respectively. Values are single-run (Molmo deterministic conditions seed~0; Qwen seed~42), except shuffled rows with seeded reruns, which are three-seed means: \texttt{shuf}, \texttt{rop-shuf}, and \texttt{shuf-corr} for Molmo (all scales) and Qwen3-VL-4B, and \texttt{shuf} for Qwen3-VL-8B; spreads in Appendix~\ref{app:seeds}. The patching appendices report their own same-run baselines, whose \texttt{base} rows differ from these tables by at most $0.007$. $n=1580$ on 4B/8B; $n=1564$ on O-7B. Conditions follow the naming in Tables~\ref{tab:conditions-main} and~\ref{tab:conditions-full}.

\begin{table}[h]
\centering
\small
\caption{Per-scale audit decomposition behind main-text Table~\ref{tab:audit}. Each cell is \emph{multi-frame gain}\,/\,\emph{order-gain}: accuracy(\texttt{base}) minus accuracy(\texttt{single-frame}), and minus accuracy(\texttt{shuf}) (three-seed mean shuffles; Supp.~\ref{app:seeds}). Family means in Table~\ref{tab:audit} average Molmo2 4B/8B/O-7B and Qwen3-VL 4B/8B.}
\label{tab:audit-perscale}
\resizebox{\columnwidth}{!}{%
\begin{tabular}{lccccc}
\toprule
\textbf{Aspect} & \textbf{Molmo 4B} & \textbf{Molmo 8B} & \textbf{Molmo O-7B} & \textbf{Qwen 4B} & \textbf{Qwen 8B} \\
\midrule
attribute\_change & $+0.45$/$+0.39$ & $+0.44$/$+0.38$ & $+0.48$/$+0.34$ & $+0.49$/$+0.35$ & $+0.55$/$+0.40$ \\
order             & $+0.29$/$+0.33$ & $+0.28$/$+0.35$ & $+0.29$/$+0.31$ & $+0.19$/$+0.27$ & $+0.23$/$+0.32$ \\
direction         & $+0.20$/$+0.16$ & $+0.15$/$+0.12$ & $+0.17$/$+0.14$ & $+0.16$/$+0.11$ & $+0.23$/$+0.13$ \\
speed             & $+0.10$/$+0.06$ & $+0.09$/$+0.04$ & $+0.14$/$+0.05$ & $+0.11$/$+0.03$ & $+0.11$/$-0.00$ \\
action            & $+0.06$/$+0.01$ & $+0.09$/$+0.02$ & $+0.12$/$+0.00$ & $+0.04$/$+0.00$ & $+0.05$/$+0.02$ \\
\midrule
Overall           & $+0.21$/$+0.18$ & $+0.20$/$+0.18$ & $+0.23$/$+0.16$ & $+0.19$/$+0.14$ & $+0.23$/$+0.16$ \\
\bottomrule
\end{tabular}}
\end{table}

\begin{table}[h]
\centering
\small
\caption{Molmo2 TempCompass overall accuracy across model sizes. Shuffled rows (\texttt{shuf}, \texttt{rop-shuf}, \texttt{shuf-corr}) are three-seed means. \texttt{base-nts}: timestamps removed. \texttt{shuf-mvts}: timestamps follow shuffled frames. \texttt{rop-shuf}: correct pixels, shuffled RoPE. \texttt{rev-mvts}: reversed pixels with timestamps moved consistently. \texttt{conf-ts}: forward pixels, reversed timestamps. \texttt{conf-shuf-ts}: pixels and timestamps shuffled with independent permutations.}
\label{tab:tempcompass-molmo}
\begin{tabular}{lccc}
\toprule
\textbf{Condition} & \textbf{4B} & \textbf{8B} & \textbf{O-7B} \\
\midrule
\texttt{base}         & 0.728 & 0.737 & 0.718 \\
\texttt{base-nts}     & 0.700 & 0.665 & 0.621 \\
\texttt{single}       & 0.514 & 0.534 & 0.484 \\
\texttt{ts-only}      & 0.412 & 0.393 & 0.406 \\
\texttt{blank-nts}    & 0.425 & 0.441 & --- \\
\midrule
\texttt{shuf}         & 0.545 & 0.561 & 0.556 \\
\texttt{shuf-mvts}    & 0.590 & 0.592 & 0.570 \\
\texttt{shuf-nts}     & 0.544 & 0.550 & 0.547 \\
\texttt{rop-shuf}     & 0.553 & 0.580 & 0.579 \\
\texttt{shuf-corr}    & 0.715 & 0.716 & 0.654 \\
\midrule
\texttt{rev}          & 0.442 & 0.448 & 0.447 \\
\texttt{rev-corr}     & 0.722 & 0.720 & 0.643 \\
\texttt{rev-mvts}     & 0.444 & 0.458 & --- \\
\midrule
\texttt{conf-ts}      & 0.717 & 0.730 & --- \\
\texttt{conf-shuf-ts} & 0.553 & 0.561 & --- \\
\bottomrule
\end{tabular}
\end{table}

\begin{table}[h]
\centering
\footnotesize
\caption{Molmo2 4B TempCompass per-aspect. Shuffled rows are three-seed means. action is at ceiling under all interventions; speed is low accuracy and non-diagnostic (Section~\ref{sec:methods:benchmarks}). Per-aspect analysis in the main text uses attribute, direction, and order. Reversal rows (\texttt{rev}, \texttt{rev-corr}, \texttt{rev-mvts}) include the paired \texttt{\_reverse} videos and are not comparable to the main-text E (Table~\ref{tab:dissoc-tc}).}
\label{tab:tempcompass-molmo4b}
\resizebox{\columnwidth}{!}{%
\begin{tabular}{lcccccc}
\toprule
\textbf{Condition} & \textbf{Overall} & \textbf{action} & \textbf{attribute} & \textbf{direction} & \textbf{order} & \textbf{speed} \\
\midrule
\texttt{base}         & 0.728 & 0.967 & 0.819 & 0.552 & 0.775 & 0.533 \\
\texttt{base-nts}     & 0.700 & 0.964 & 0.760 & 0.537 & 0.742 & 0.495 \\
\texttt{single}       & 0.514 & 0.908 & 0.368 & 0.349 & 0.480 & 0.432 \\
\texttt{ts-only}      & 0.412 & 0.497 & 0.389 & 0.376 & 0.387 & 0.404 \\
\texttt{blank-nts}    & 0.425 & 0.512 & 0.420 & 0.367 & 0.401 & 0.420 \\
\midrule
\texttt{shuf}         & 0.545 & 0.960 & 0.426 & 0.391 & 0.443 & 0.469 \\
\texttt{shuf-mvts}    & 0.590 & 0.962 & 0.563 & 0.417 & 0.516 & 0.469 \\
\texttt{shuf-nts}     & 0.544 & 0.956 & 0.450 & 0.396 & 0.432 & 0.452 \\
\texttt{rop-shuf}     & 0.553 & 0.960 & 0.431 & 0.402 & 0.463 & 0.473 \\
\texttt{shuf-corr}    & 0.715 & 0.965 & 0.812 & 0.541 & 0.756 & 0.503 \\
\midrule
\texttt{rev}          & 0.442 & 0.964 & 0.153 & 0.337 & 0.149 & 0.539 \\
\texttt{rev-corr}     & 0.722 & 0.967 & 0.809 & 0.540 & 0.755 & 0.543 \\
\texttt{rev-mvts}     & 0.444 & 0.959 & 0.174 & 0.331 & 0.169 & 0.521 \\
\midrule
\texttt{conf-ts}      & 0.717 & 0.967 & 0.788 & 0.546 & 0.745 & 0.539 \\
\texttt{conf-shuf-ts} & 0.553 & 0.964 & 0.469 & 0.394 & 0.444 & 0.463 \\
\bottomrule
\end{tabular}
}
\end{table}

\begin{table}[h]
\centering
\footnotesize
\caption{Qwen3-VL 4B TempCompass per-aspect. \texttt{shuf-full}: shuffle with both timestamps moved and all RoPE axes corrected. \texttt{rop-shuf}/\texttt{rop-rev}: see Table~\ref{tab:conditions-full}. \texttt{rev-mvts}: pixels and timestamps consistently reversed. Reversal rows (\texttt{rev}, \texttt{rev-corr}, \texttt{rev-mvts}) include the paired \texttt{\_reverse} videos and are not comparable to the main-text E (Table~\ref{tab:dissoc-tc}).}
\label{tab:tempcompass-qwen4b}
\resizebox{\columnwidth}{!}{%
\begin{tabular}{lcccccc}
\toprule
\textbf{Condition} & \textbf{Overall} & \textbf{action} & \textbf{attribute} & \textbf{direction} & \textbf{order} & \textbf{speed} \\
\midrule
\texttt{base}        & 0.710 & 0.970 & 0.781 & 0.504 & 0.768 & 0.530 \\
\texttt{base-nts}    & 0.696 & 0.976 & 0.778 & 0.487 & 0.738 & 0.505 \\
\texttt{single}      & 0.522 & 0.932 & 0.288 & 0.349 & 0.583 & 0.423 \\
\texttt{ts-only}     & 0.384 & 0.476 & 0.340 & 0.313 & 0.434 & 0.353 \\
\texttt{blank-nts}   & 0.392 & 0.538 & 0.330 & 0.334 & 0.411 & 0.338 \\
\midrule
\texttt{shuf}        & 0.566 & 0.970 & 0.434 & 0.397 & 0.496 & 0.498 \\
\texttt{shuf-nts}    & 0.561 & 0.967 & 0.451 & 0.415 & 0.447 & 0.492 \\
\texttt{shuf-mvts}   & 0.618 & 0.976 & 0.618 & 0.439 & 0.530 & 0.508 \\
\texttt{shuf-corr}   & 0.611 & 0.969 & 0.584 & 0.435 & 0.549 & 0.498 \\
\texttt{shuf-full}   & 0.605 & 0.976 & 0.583 & 0.427 & 0.507 & 0.511 \\
\texttt{rop-shuf}    & 0.671 & 0.973 & 0.676 & 0.466 & 0.715 & 0.518 \\
\texttt{rop-rev}     & 0.648 & 0.973 & 0.611 & 0.442 & 0.699 & 0.505 \\
\midrule
\texttt{conf-ts}     & 0.659 & 0.972 & 0.635 & 0.482 & 0.698 & 0.499 \\
\texttt{rev}         & 0.455 & 0.970 & 0.194 & 0.334 & 0.209 & 0.505 \\
\texttt{rev-corr}    & 0.493 & 0.964 & 0.337 & 0.379 & 0.219 & 0.514 \\
\texttt{rev-mvts}    & 0.492 & 0.971 & 0.342 & 0.350 & 0.256 & 0.490 \\
\texttt{conf-ts2}    & 0.565 & 0.969 & 0.439 & 0.388 & 0.489 & 0.506 \\
\bottomrule
\end{tabular}
}
\end{table}

\begin{table}[h]
\centering
\footnotesize
\caption{Qwen3-VL 8B TempCompass per-aspect. Conditions and column abbreviations as in Table~\ref{tab:tempcompass-qwen4b}. Reversal rows (\texttt{rev}, \texttt{rev-corr}, \texttt{rev-mvts}) include the paired \texttt{\_reverse} videos and are not comparable to the main-text E (Table~\ref{tab:dissoc-tc}).}
\label{tab:tempcompass-qwen8b}
\resizebox{\columnwidth}{!}{%
\begin{tabular}{lcccccc}
\toprule
\textbf{Condition} & \textbf{Overall} & \textbf{action} & \textbf{attribute} & \textbf{direction} & \textbf{order} & \textbf{speed} \\
\midrule
\texttt{base}        & 0.743 & 0.976 & 0.830 & 0.528 & 0.808 & 0.580 \\
\texttt{base-nts}    & 0.704 & 0.973 & 0.788 & 0.499 & 0.745 & 0.521 \\
\texttt{single}      & 0.518 & 0.920 & 0.285 & 0.313 & 0.556 & 0.479 \\
\texttt{ts-only}     & 0.406 & 0.574 & 0.330 & 0.319 & 0.407 & 0.388 \\
\texttt{blank-nts}   & 0.376 & 0.524 & 0.309 & 0.290 & 0.404 & 0.344 \\
\midrule
\texttt{shuf}        & 0.579 & 0.960 & 0.428 & 0.400 & 0.490 & 0.584 \\
\texttt{shuf-nts}    & 0.558 & 0.959 & 0.431 & 0.418 & 0.424 & 0.521 \\
\texttt{shuf-mvts}   & 0.651 & 0.967 & 0.674 & 0.445 & 0.599 & 0.562 \\
\texttt{shuf-corr}   & 0.628 & 0.967 & 0.656 & 0.445 & 0.513 & 0.543 \\
\texttt{shuf-full}   & 0.627 & 0.967 & 0.632 & 0.439 & 0.530 & 0.549 \\
\texttt{rop-shuf}    & 0.708 & 0.973 & 0.757 & 0.484 & 0.748 & 0.580 \\
\texttt{rop-rev}     & 0.687 & 0.979 & 0.667 & 0.469 & 0.719 & 0.596 \\
\midrule
\texttt{conf-ts}     & 0.723 & 0.970 & 0.789 & 0.517 & 0.784 & 0.558 \\
\texttt{rev}         & 0.452 & 0.959 & 0.163 & 0.319 & 0.169 & 0.584 \\
\texttt{rev-corr}    & 0.498 & 0.967 & 0.330 & 0.346 & 0.228 & 0.568 \\
\texttt{rev-mvts}    & 0.462 & 0.964 & 0.199 & 0.320 & 0.184 & 0.581 \\
\texttt{conf-ts2}    & 0.586 & 0.963 & 0.450 & 0.409 & 0.511 & 0.566 \\
\bottomrule
\end{tabular}
}
\end{table}

% ------------------------------------------------------------
\section{Architecture Dissociation Under Reversal}
\noindent\textit{Supports Sec.~\ref{sec:behavior:rope}: full per-aspect $E$/$F$ tables, the mirror control, and the motion-dependent/static split behind main-text Table~\ref{tab:reversal-tc}.}
\label{app:dissociation}
% ------------------------------------------------------------

Under \texttt{rev-corr} (reversed pixels, RoPE corrected to forward order) on the original videos, the
two families resolve the pixel-vs-RoPE conflict through opposite channels. Table~\ref{tab:dissoc-tc}
scores against the forward ground truth: Molmo's accuracy is preserved (\texttt{rev-corr}~$\approx$~\texttt{base}),
so it reads order from RoPE, while Qwen's collapses, so it reads order from the pixels. The
\emph{mirror} column applies the opposite conflict (forward pixels, reversed RoPE; the paired
\texttt{\_reverse} videos under \texttt{rev-corr}) graded against the forward content shown: the
pattern flips, Molmo now fails and Qwen succeeds, confirming that Molmo follows RoPE and Qwen follows
the pixels in \emph{both} conflict directions. (\texttt{rev-corr} here is original-video accuracy and
equals the main-text E, Table~\ref{tab:reversal-fe}, up to the few non-informative pairs.) TempCompass permutes the answer options between \texttt{\{id\}} and \texttt{\{id\}\_reverse}, so the reversed GT is matched by option \emph{text} rather than letter, and we restrict to the informative pairs whose original and reversed GT differ (a few pairs per aspect whose answer is unchanged by reversal are excluded). All scoring uses one shared text-normalization policy (punctuation-stripping), and rows whose reversed-GT text is not among the item's options are excluded rather than scored, since they could never count as read-reversed; on \emph{direction} this removes 5 rows per family and shifts values at the third decimal relative to round 1. The \emph{direction} counts chain as: 157 reverse items; 153 string-matched pairs (four questions rewritten under reversal, Supp.~\ref{app:revclip}); 147 (Molmo, index pairing) or 144 (Qwen, stem pairing) informative pairs with the reversed label offered.

\begin{table}[h]
\centering
\small
\caption{Architecture dissociation under reversal (TempCompass, original videos). \texttt{base}:
forward pixels and RoPE. \texttt{rev-corr}: reversed pixels, forward RoPE (graded vs forward GT).
\emph{mirror}: forward pixels, reversed RoPE on the paired \texttt{\_reverse} videos (graded vs the
forward content shown). Molmo preserves accuracy under \texttt{rev-corr} and fails under \emph{mirror}
(follows RoPE); Qwen does the reverse (follows pixels).}
\label{tab:dissoc-tc}
\resizebox{\columnwidth}{!}{%
\begin{tabular}{lccc|ccc}
\toprule
& \multicolumn{3}{c|}{\emph{attribute\_change}} & \multicolumn{3}{c}{\emph{direction}} \\
\textbf{Model} & base & rev-corr & mirror & base & rev-corr & mirror \\
\midrule
Molmo2 4B   & 0.965 & 0.965 & 0.312 & 0.640 & 0.635 & 0.401 \\
Molmo2 8B   & 0.972 & 0.965 & 0.368 & 0.617 & 0.611 & 0.396 \\
Molmo2 O-7B & 0.965 & 0.882 & 0.458 & 0.607 & 0.573 & 0.471 \\
Qwen3-VL 4B & 0.944 & 0.576 & 0.868 & 0.657 & 0.511 & 0.569 \\
Qwen3-VL 8B & 0.972 & 0.549 & 0.896 & 0.596 & 0.449 & 0.608 \\
\bottomrule
\end{tabular}}
\end{table}

On TVBench (no paired \texttt{\_reverse} videos, so \texttt{rev-corr} is applied directly to the
clips), the same dissociation appears across temporal tasks (main-text Table~\ref{tab:dissoc-tvb}):
Molmo's accuracy is essentially unchanged by reversal at both scales (mean temporal-task drop $+0.03$),
while Qwen's collapses on every temporal task at both scales (Qwen-8B overall $0.552 \rightarrow 0.335$ (micro-averaged)).

Restricting the TempCompass reversal to the motion-dependent slice (correct from the full video, wrong
from a single frame) confirms the override is not an artifact of statically answerable items
(Table~\ref{tab:reversal-md}). Molmo's override is as strong or stronger on the slice, while Qwen's
(4B and 8B, \emph{attribute\_change}) is confined to the static \emph{rest} and vanishes where motion
is required. At both scales the motion-dependent slice sits at or above zero while the static rest
stays strongly negative, with the two CIs non-overlapping at 8B. The paired bootstrap CIs are Qwen
\emph{attribute\_change} motion-dependent slice 4B $+0.01$ $[-0.21,+0.23]$ and 8B $+0.24$
$[+0.01,+0.45]$, against a strongly negative static rest (4B $-0.49$, 8B $-0.57$).

Qwen \emph{direction} does not dissociate at either scale, and its rows in
Table~\ref{tab:reversal-md} show why it cannot. Reversed direction content sits at Qwen's
trackability floor with no intervention at all (reversed-clip base accuracy $0.327$/$0.444$,
Appendix~\ref{app:revclip}), and $F-E$ is a single condition rather than a contrast, so the shared
forward prior alone drags it negative for any model that cannot track the reversed content visually.
Qwen's motion-dependent slice equals its static rest within noise ($-0.235$ $[-0.53,+0.06]$ vs
$-0.255$ at 4B, $-0.130$ $[-0.39,+0.13]$ vs $-0.061$ at 8B): no positional differentiation on
precisely the items where position use would show. Molmo's override instead concentrates exactly
there ($-0.95$ on the slice vs $-0.23$ on the rest). Position reading needs no content trackability,
which is why Molmo's signal survives the floor that silences Qwen's. \emph{direction} is thus a
second instance of the scope condition stated for \emph{action\_sequence}: a deterministic-flip task
screens the channel only where reversed content is trackable.

Tables~\ref{tab:reversal-fe} and~\ref{tab:reversal-md} give the full per-aspect breakdown that
main-text Table~\ref{tab:reversal-tc} consolidates. Table~\ref{tab:reversal-fe} reports the
original-content ($E$) and reversed-content ($F$) accuracies for both diagnostic aspects, and
Table~\ref{tab:reversal-md} splits $F-E$ into the motion-dependent slice and the static rest.

\begin{table}[t]
\centering
\small
\caption{Reversal override (full per-aspect): \texttt{rev-corr} (reversed pixels, corrected RoPE)
scored against original GT (E) and reversed GT (F). $F-E < 0$ indicates order read from RoPE rather
than the pixels. F uses option-text alignment (TempCompass permutes option letters under reversal),
informative items only (original GT $\neq$ reversed GT). The \emph{attribute\_change} $F-E$ column is
carried in main-text Table~\ref{tab:reversal-tc}.}
\label{tab:reversal-fe}
\resizebox{\columnwidth}{!}{%
\begin{tabular}{lcccccc}
\toprule
\textbf{Model} & \multicolumn{3}{c}{\emph{attribute\_change}} & \multicolumn{3}{c}{\emph{direction}} \\
& E & F & $F-E$ & E & F & $F-E$ \\
\midrule
Molmo2 4B   & 0.972 & 0.014 & $-0.957$ & 0.633 & 0.224 & $-0.408$ \\
Molmo2 8B   & 0.965 & 0.014 & $-0.950$ & 0.618 & 0.236 & $-0.382$ \\
Molmo2 O-7B & 0.894 & 0.078 & $-0.816$ & 0.565 & 0.293 & $-0.272$ \\
Qwen3-VL 4B & 0.582 & 0.348 & $-0.234$ & 0.528 & 0.278 & $-0.250$ \\
Qwen3-VL 8B & 0.546 & 0.411 & $-0.135$ & 0.444 & 0.361 & $-0.083$ \\
\bottomrule
\end{tabular}
}
\end{table}

\begin{table}[t]
\centering
\small
\caption{Reversal override on the motion-dependent slice (correct from the full video, wrong from a
single frame). The \emph{full} column is the $F-E$ of Table~\ref{tab:reversal-fe}, split here into the
motion-dependent slice (m-dep) and the static rest. $F-E<0$ is order read from RoPE, $n$ in
parentheses. Molmo holds or strengthens on the slice. Qwen (4B and 8B) \emph{attribute\_change} is
confined to the static \emph{rest}, with the motion-dependent slice at or above zero.}
\label{tab:reversal-md}
\resizebox{\columnwidth}{!}{%
\begin{tabular}{llccc}
\toprule
\textbf{Model} & \textbf{Aspect} & \textbf{full $F\!-\!E$} & \textbf{m-dep $F\!-\!E$ ($n$)} & \textbf{rest $F\!-\!E$ ($n$)} \\
\midrule
Molmo2 4B   & attr.\ change & $-0.957$ & $-1.000$\,(67) & $-0.919$\,(74) \\
Molmo2 8B   & attr.\ change & $-0.950$ & $-0.983$\,(59) & $-0.927$\,(82) \\
Molmo2 O-7B & attr.\ change & $-0.816$ & $-0.766$\,(77) & $-0.875$\,(64) \\
Molmo2 4B   & direction & $-0.408$ & $-0.946$\,(37) & $-0.227$\,(110) \\
Molmo2 8B   & direction & $-0.382$ & $-0.941$\,(34) & $-0.209$\,(110) \\
Molmo2 O-7B & direction & $-0.272$ & $-0.698$\,(43) & $-0.096$\,(104) \\
\midrule
Qwen3-VL 4B & attr.\ change & $-0.234$ & $+0.014$\,(71) & $-0.486$\,(70) \\
Qwen3-VL 8B & attr.\ change & $-0.135$ & $+0.237$\,(76) & $-0.569$\,(65) \\
Qwen3-VL 4B & direction & $-0.250$ & $-0.235$\,(34) & $-0.255$\,(110) \\
Qwen3-VL 8B & direction & $-0.083$ & $-0.130$\,(46) & $-0.061$\,(98) \\
\bottomrule
\end{tabular}}
\end{table}

% ------------------------------------------------------------
\section{Dose-Response: Multi-Frame Gain Is Not a Single-Frame Artifact}
\label{app:doseresponse}
% ------------------------------------------------------------

Multi-frame gain compares the full video to a single repeated frame, and that frame was frame~0. On
sequence tasks frame~0 can leak which event comes first, which would inflate single-frame accuracy and
understate the gain, or make the gain depend on which frame we picked. To rule this out we measure
accuracy as a function of the number of frames the model sees, $K\in\{1,2,4,8,\text{full}\}$, sampling
$K$ evenly-spaced frames with timestamps kept. Crucially, the $K{=}1$ point uses a \emph{middle} frame,
not frame~0, so it is a single-frame baseline independent of the frame-0 one. We run Molmo2-4B and
Qwen3-VL-4B on all nine TVBench tasks and pool the five temporal tasks into a model-specific
motion-dependent slice (correct from the full video, wrong from frame~0) and the static remainder
(Table~\ref{tab:doseresponse}).

\begin{table}[t]
\centering
\small
\resizebox{\columnwidth}{!}{%
\begin{tabular}{llccccccc}
\toprule
\textbf{Group} & \textbf{Model} & \boldmath$n$ & \textbf{frame0} & \textbf{$K{=}1$mid} & \textbf{$K{=}2$} & \textbf{$K{=}4$} & \textbf{$K{=}8$} & \textbf{full} \\
\midrule
multi-frame-dep & Molmo2-4B   & 396 & 0.00 & 0.18 & 0.41 & 0.60 & 0.80 & 1.00 \\
multi-frame-dep & Qwen3-VL-4B & 355 & 0.00 & 0.41 & 0.61 & 0.74 & 0.87 & 1.00 \\
\midrule
static          & Molmo2-4B   & 717 & 0.71 & 0.52 & 0.62 & 0.60 & 0.62 & 0.63 \\
static          & Qwen3-VL-4B & 758 & 0.61 & 0.50 & 0.59 & 0.56 & 0.56 & 0.54 \\
\bottomrule
\end{tabular}}
\caption{Dose-response on the TVBench temporal tasks: accuracy as the number of sampled frames $K$
grows, with $K{=}1$ a \emph{middle} frame (not frame~0). The motion-dependent slice (full-correct,
frame-0-wrong, so $\text{full}{=}1.00$ by construction) climbs monotonically with $K$, while the static
remainder stays flat from $K{=}1$ on. Molmo2-4B and Qwen3-VL-4B.}
\label{tab:doseresponse}
\end{table}

The motion-dependent items climb steeply and monotonically with $K$ (Molmo $0.18\!\to\!1.0$, Qwen
$0.41\!\to\!1.0$) while the static items stay flat at $0.5$ to $0.6$. Three points follow. First, the
climb is not a frame-0 leak: at $K{=}1$ the frame is a middle frame, independent of the frame-0
baseline, and accuracy is still near the floor there, so the gain is caused by \emph{adding} frames,
the definition of cross-frame dependence. Second, a smooth monotonic curve with a flat static control
is far harder to attribute to a baseline artifact than a two-point gap. Third, the signature replicates
across a bidirectional (Molmo) and a causal (Qwen) model, so it is a property of the tasks, not one
model's quirk. Both families are genuinely motion-dependent (the task question), which is the premise the
channel dissociation (the channel question) builds on.

% ------------------------------------------------------------
\section{TVBench Results}
\noindent\textit{Supports Secs.~\ref{sec:behavior:rope} and~\ref{sec:behavior:patching}: per-category accuracies (base columns for main-text Table~\ref{tab:dissoc-tvb}) and the TVBench patching replication (Qwen3-VL, \emph{moving\_direction}, Table~\ref{tab:tvbench-patch}).}
\label{app:tvbench}
% ------------------------------------------------------------

Tables~\ref{tab:tvbench-molmo}--\ref{tab:tvbench-qwen8b} report TVBench accuracy for all models and conditions. Macro-average is computed with equal weight per category. Chance baselines vary by task (0.500 for 2-choice tasks, 0.333 for 3-choice, 0.250 for 4-choice); the reported macro-average chance level is 0.315.

Table~\ref{tab:dissoc-tvb} gives the per-task reversal-drop screen of
Section~\ref{sec:behavior:rope} with paired bootstrap 95\% CIs (10{,}000
resamples over items, base and \texttt{rev-corr} paired per item).
Reversal is deterministic, so the CIs quantify item-sampling stability,
not run-to-run variance. Every Qwen temporal-task drop excludes zero;
Molmo's drops distinguishable from zero are \emph{moving\_direction} 4B
($+0.08$), \emph{action\_localization} 8B ($+0.06$), and
\emph{egocentric\_sequence} 8B ($+0.14$). Both families are near zero on
the non-temporal tasks, so the conflict's cost is specific to temporal
content.

\begin{table}[h]
\centering
\small
\caption{The label-free screen: per-task TVBench reversal-drop,
accuracy(\texttt{base}) minus accuracy(\texttt{rev-corr}) against forward
labels, with paired bootstrap 95\% CIs. $n$ is paired items per task
(Qwen-8B pairs one fewer on \emph{action\_sequence} and
\emph{unexpected\_action}). \texttt{object\_shuffle} is non-temporal.
Base accuracies in Tables~\ref{tab:tvbench-molmo}--\ref{tab:tvbench-qwen8b}.}
\label{tab:dissoc-tvb}
\resizebox{\columnwidth}{!}{%
\begin{tabular}{lccccc}
\toprule
\textbf{Task} & $n$ & \textbf{Molmo 4B} & \textbf{Molmo 8B} & \textbf{Qwen 4B} & \textbf{Qwen 8B} \\
\midrule
scene\_transition    & 185 & $+0.01$ $[-0.02,+0.04]$ & $+0.00$ $[-0.03,+0.03]$ & $+0.55$ $[+0.46,+0.64]$ & $+0.55$ $[+0.46,+0.64]$ \\
moving\_direction    & 232 & $+0.08$ $[+0.03,+0.14]$ & $+0.02$ $[-0.03,+0.07]$ & $+0.41$ $[+0.34,+0.48]$ & $+0.33$ $[+0.27,+0.40]$ \\
action\_sequence     & 437 & $+0.00$ $[-0.01,+0.02]$ & $+0.01$ $[-0.01,+0.03]$ & $+0.36$ $[+0.30,+0.42]$ & $+0.32$ $[+0.26,+0.38]$ \\
action\_localization & 160 & $+0.02$ $[-0.02,+0.06]$ & $+0.06$ $[+0.01,+0.12]$ & $+0.29$ $[+0.21,+0.38]$ & $+0.28$ $[+0.19,+0.37]$ \\
egocentric\_sequence & 200 & $-0.01$ $[-0.03,+0.02]$ & $+0.14$ $[+0.09,+0.18]$ & $+0.24$ $[+0.17,+0.32]$ & $+0.28$ $[+0.21,+0.35]$ \\
object\_count        & 148 & $+0.01$ $[-0.05,+0.06]$ & $+0.03$ $[-0.02,+0.09]$ & $+0.18$ $[+0.09,+0.27]$ & $+0.09$ $[+0.01,+0.16]$ \\
unexpected\_action   &  82 & $+0.04$ $[-0.02,+0.10]$ & $+0.06$ $[+0.00,+0.12]$ & $+0.11$ $[+0.00,+0.22]$ & $+0.10$ $[-0.02,+0.22]$ \\
action\_count        & 536 & $+0.00$ $[-0.03,+0.03]$ & $-0.00$ $[-0.03,+0.02]$ & $+0.06$ $[+0.02,+0.10]$ & $+0.08$ $[+0.04,+0.12]$ \\
object\_shuffle      & 225 & $-0.01$ $[-0.06,+0.03]$ & $-0.06$ $[-0.12,+0.00]$ & $-0.05$ $[-0.12,+0.01]$ & $-0.02$ $[-0.07,+0.04]$ \\
\bottomrule
\end{tabular}}
\end{table}

\begin{table}[h]
\centering
\small
\caption{Molmo2 TVBench macro-average accuracy. \texttt{shuf} and \texttt{rop-shuf} rows are three-seed means (Supp.~\ref{app:seeds}); other rows single-run. \texttt{shuf-corr} $\approx$ \texttt{base} across all sizes, replicating the TempCompass recovery pattern. \texttt{shuf-mvts} $>$ \texttt{shuf} replicates the timestamp-moving benefit.}
\label{tab:tvbench-molmo}
\begin{tabular}{lccc}
\toprule
\textbf{Condition} & \textbf{4B} & \textbf{8B} & \textbf{O-7B} \\
\midrule
chance            & 0.315 & 0.315 & 0.315 \\
\texttt{base}     & 0.662 & 0.647 & 0.552 \\
\texttt{single}   & 0.355 & 0.356 & 0.332 \\
\texttt{shuf}     & 0.385 & 0.379 & 0.362 \\
\texttt{shuf-mvts} & 0.447 & 0.434 & 0.384 \\
\texttt{shuf-nts}  & 0.396 & 0.397 & 0.377 \\
\texttt{rop-shuf}  & 0.402 & 0.408 & 0.395 \\
\texttt{shuf-corr} & 0.657 & 0.631 & 0.522 \\
\midrule
\texttt{shuf-corr} $-$ \texttt{base} & $-0.005$ & $-0.016$ & $-0.030$ \\
\texttt{shuf-mvts} $-$ \texttt{shuf} & $+0.062$ & $+0.055$ & $+0.022$ \\
\bottomrule
\end{tabular}
\end{table}

\begin{table*}[t]
\centering
\footnotesize
\caption{Molmo2 4B TVBench per-task accuracy. ac=action\_count, al=action\_localization, as=action\_sequence, es=egocentric\_sequence, md=moving\_direction, oc=object\_count, os=object\_shuffle, st=scene\_transition, ua=unexpected\_action. \texttt{shuf} and \texttt{rop-shuf} rows are three-seed means.}
\label{tab:tvbench-molmo4b-pertask}
\begin{tabular}{lcccccccccc}
\toprule
\textbf{Cond} & \textbf{Macro} & \textbf{ac} & \textbf{al} & \textbf{as} & \textbf{es} & \textbf{md} & \textbf{oc} & \textbf{os} & \textbf{st} & \textbf{ua} \\
\midrule
chance & 0.315 & 0.250 & 0.250 & 0.500 & 0.250 & 0.250 & 0.250 & 0.333 & 0.500 & 0.250 \\
\texttt{base}      & 0.662 & 0.556 & 0.600 & 0.826 & 0.720 & 0.703 & 0.757 & 0.373 & 0.886 & 0.537 \\
\texttt{base-nts}  & 0.610 & 0.507 & 0.506 & 0.808 & 0.685 & 0.724 & 0.486 & 0.396 & 0.886 & 0.488 \\
\texttt{shuf-corr} & 0.657 & 0.558 & 0.606 & 0.817 & 0.730 & 0.681 & 0.730 & 0.382 & 0.881 & 0.532 \\
\texttt{shuf}      & 0.385 & 0.394 & 0.421 & 0.515 & 0.265 & 0.277 & 0.338 & 0.336 & 0.586 & 0.329 \\
\texttt{shuf-mvts} & 0.447 & 0.420 & 0.419 & 0.609 & 0.365 & 0.431 & 0.331 & 0.338 & 0.719 & 0.392 \\
\texttt{shuf-nts}  & 0.396 & 0.379 & 0.419 & 0.535 & 0.365 & 0.250 & 0.304 & 0.382 & 0.584 & 0.342 \\
\texttt{rop-shuf}  & 0.402 & 0.394 & 0.396 & 0.547 & 0.305 & 0.330 & 0.356 & 0.344 & 0.634 & 0.313 \\
\texttt{shuf-full} & 0.638 & 0.554 & 0.600 & 0.815 & 0.720 & 0.664 & 0.669 & 0.342 & 0.865 & 0.512 \\
\texttt{rev-corr}  & 0.646 & 0.554 & 0.581 & 0.824 & 0.725 & 0.621 & 0.750 & 0.387 & 0.876 & 0.500 \\
\texttt{conf-ts}   & 0.639 & 0.582 & 0.562 & 0.808 & 0.690 & 0.681 & 0.655 & 0.369 & 0.876 & 0.524 \\
\bottomrule
\end{tabular}
\end{table*}

\begin{table*}[t]
\centering
\footnotesize
\caption{Qwen3-VL 4B TVBench per-task accuracy. Same column abbreviations as Table~\ref{tab:tvbench-molmo4b-pertask}. \texttt{shuf}, \texttt{shuf-corr}, and \texttt{rop-shuf} rows are three-seed means (shared permutations); other shuffle variants are single-run. \texttt{shuf-corr} $\not\approx$ \texttt{base} (gap $-0.097$), consistent with TempCompass.}
\label{tab:tvbench-qwen4b}
\begin{tabular}{lcccccccccc}
\toprule
\textbf{Cond} & \textbf{Macro} & \textbf{ac} & \textbf{al} & \textbf{as} & \textbf{es} & \textbf{md} & \textbf{oc} & \textbf{os} & \textbf{st} & \textbf{ua} \\
\midrule
chance             & 0.315 & 0.250 & 0.250 & 0.500 & 0.250 & 0.250 & 0.250 & 0.333 & 0.500 & 0.250 \\
\texttt{base}      & 0.565 & 0.368 & 0.562 & 0.810 & 0.645 & 0.470 & 0.480 & 0.360 & 0.865 & 0.524 \\
\texttt{base-nts}  & 0.577 & 0.371 & 0.569 & 0.808 & 0.665 & 0.547 & 0.480 & 0.373 & 0.870 & 0.512 \\
\texttt{base-nds}  & 0.540 & 0.381 & 0.537 & 0.808 & 0.710 & 0.332 & 0.358 & 0.364 & 0.849 & 0.524 \\
\texttt{shuf}      & 0.400 & 0.320 & 0.327 & 0.614 & 0.407 & 0.191 & 0.349 & 0.410 & 0.551 & 0.435 \\
\texttt{shuf-mvts} & 0.463 & 0.317 & 0.375 & 0.650 & 0.445 & 0.427 & 0.412 & 0.360 & 0.622 & 0.561 \\
\texttt{shuf-corr}    & 0.468 & 0.320 & 0.435 & 0.693 & 0.477 & 0.239 & 0.468 & 0.403 & 0.670 & 0.504 \\
\texttt{shuf-corr-T}  & 0.409 & 0.321 & 0.356 & 0.638 & 0.410 & 0.181 & 0.358 & 0.387 & 0.578 & 0.451 \\
\texttt{shuf-corr-HW} & 0.440 & 0.326 & 0.419 & 0.659 & 0.470 & 0.190 & 0.426 & 0.400 & 0.611 & 0.463 \\
\texttt{shuf-corr-ds} & 0.484 & 0.315 & 0.456 & 0.684 & 0.460 & 0.349 & 0.439 & 0.440 & 0.714 & 0.500 \\
\texttt{shuf-full}    & 0.468 & 0.302 & 0.431 & 0.693 & 0.400 & 0.349 & 0.439 & 0.427 & 0.659 & 0.512 \\
\texttt{shuf-nds}     & 0.409 & 0.336 & 0.344 & 0.604 & 0.470 & 0.228 & 0.338 & 0.400 & 0.562 & 0.402 \\
\texttt{shuf-corr-nds} & 0.475 & 0.310 & 0.450 & 0.712 & 0.535 & 0.284 & 0.419 & 0.391 & 0.686 & 0.488 \\
\texttt{rop-shuf}     & 0.537 & 0.328 & 0.479 & 0.778 & 0.515 & 0.569 & 0.423 & 0.416 & 0.825 & 0.500 \\
\texttt{rop-rev}      & 0.508 & 0.269 & 0.469 & 0.771 & 0.485 & 0.616 & 0.324 & 0.387 & 0.773 & 0.476 \\
\midrule
\texttt{rev}          & 0.302 & 0.371 & 0.200 & 0.426 & 0.340 & 0.065 & 0.270 & 0.413 & 0.227 & 0.402 \\
\texttt{rev-corr}     & 0.325 & 0.304 & 0.269 & 0.453 & 0.400 & 0.056 & 0.297 & 0.413 & 0.314 & 0.415 \\
\bottomrule
\end{tabular}
\end{table*}

\begin{table*}[t]
\centering
\footnotesize
\caption{Qwen3-VL 8B TVBench per-task accuracy. \texttt{shuf}, \texttt{shuf-corr}, and \texttt{rop-shuf} rows are three-seed means (shared permutations); \texttt{shuf-corr} and \texttt{shuf} are statistically tied on moving\_direction (see TVBench notes). \texttt{shuf-mvts} $>$ \texttt{shuf-corr} on moving\_direction ($+0.261$), consistent with timestamps dominating over RoPE on that task.}
\label{tab:tvbench-qwen8b}
\begin{tabular}{lcccccccccc}
\toprule
\textbf{Cond} & \textbf{Macro} & \textbf{ac} & \textbf{al} & \textbf{as} & \textbf{es} & \textbf{md} & \textbf{oc} & \textbf{os} & \textbf{st} & \textbf{ua} \\
\midrule
chance             & 0.315 & 0.250 & 0.250 & 0.500 & 0.250 & 0.250 & 0.250 & 0.333 & 0.500 & 0.250 \\
\texttt{base}      & 0.558 & 0.381 & 0.544 & 0.814 & 0.685 & 0.379 & 0.392 & 0.324 & 0.870 & 0.630 \\
\texttt{base-nts}  & 0.533 & 0.375 & 0.494 & 0.823 & 0.670 & 0.297 & 0.365 & 0.360 & 0.903 & 0.506 \\
\texttt{base-nds}  & 0.547 & 0.379 & 0.519 & 0.826 & 0.775 & 0.323 & 0.338 & 0.347 & 0.865 & 0.556 \\
\texttt{shuf}      & 0.373 & 0.243 & 0.390 & 0.611 & 0.452 & 0.083 & 0.351 & 0.360 & 0.537 & 0.329 \\
\texttt{shuf-mvts} & 0.444 & 0.315 & 0.431 & 0.647 & 0.455 & 0.341 & 0.338 & 0.324 & 0.649 & 0.494 \\
\texttt{shuf-corr}    & 0.437 & 0.289 & 0.427 & 0.687 & 0.538 & 0.080 & 0.322 & 0.364 & 0.661 & 0.564 \\
\texttt{shuf-corr-T}  & 0.381 & 0.267 & 0.388 & 0.606 & 0.475 & 0.069 & 0.345 & 0.316 & 0.557 & 0.407 \\
\texttt{shuf-corr-HW} & 0.415 & 0.295 & 0.362 & 0.642 & 0.495 & 0.103 & 0.318 & 0.396 & 0.595 & 0.531 \\
\texttt{shuf-corr-ds} & 0.457 & 0.299 & 0.450 & 0.700 & 0.515 & 0.159 & 0.351 & 0.382 & 0.714 & 0.543 \\
\texttt{shuf-full}    & 0.429 & 0.317 & 0.419 & 0.656 & 0.505 & 0.129 & 0.297 & 0.338 & 0.659 & 0.543 \\
\texttt{shuf-nds}     & 0.369 & 0.263 & 0.381 & 0.606 & 0.455 & 0.159 & 0.297 & 0.333 & 0.519 & 0.309 \\
\texttt{shuf-corr-nds} & 0.443 & 0.287 & 0.400 & 0.711 & 0.585 & 0.112 & 0.331 & 0.284 & 0.735 & 0.543 \\
\texttt{rop-shuf}     & 0.503 & 0.260 & 0.485 & 0.782 & 0.540 & 0.405 & 0.374 & 0.351 & 0.834 & 0.494 \\
\bottomrule
\end{tabular}
\end{table*}

\paragraph{TVBench notes.}
Reversal conditions (\texttt{rev}, \texttt{rev-corr}, \texttt{rev-mvts}) for the three deterministic-flip tasks are reported with two-label ($E$/$F$) scoring at both scales in Appendix~\ref{app:revgt}, which supersedes forward-label-only readings of those rows.
The main patterns from TempCompass replicate: \texttt{shuf-corr} $\approx$ \texttt{base} (Molmo), \texttt{shuf-corr} $\not\approx$ \texttt{base} (Qwen), \texttt{shuf-mvts} $>$ \texttt{shuf} (timestamps help when moved with frames). On moving\_direction, the shared-permutation three-seed means give \texttt{shuf-corr} $>$ \texttt{shuf} at 4B (0.239 vs 0.191) and a statistical tie at 8B (0.080 vs 0.083, within 1~SE at $n=232$); earlier single-draw runs with independent permutations per condition showed inverted orderings on this task, a sampling artifact the seeded means replace. Activation patching (\texttt{act\_ds}) fully recovers moving\_direction at both scales (4B: act\_ds $= 0.487 \approx$ q0 $= 0.461$, $\Delta = +0.026$; 8B: act\_ds $= 0.388 \approx$ q0 $= 0.397$, $\Delta = -0.009$; both within noise). act\_early $=$ act\_ds at both scales, confirming saturation at the DeepStack injection depth (layers~0--2). Act\_layer0 residuals ($-0.121$/$-0.099$ at 4B/8B) are substantially larger than on TempCompass overall ($-0.019$/$-0.007$), consistent with motion-direction answers being more directly dependent on visual content order. Conv3d baking does not contribute at either scale (e1 $\approx$ \texttt{shuf}: $+0.013$ at 4B, $-0.009$ at 8B). One unexplained observation remains: \texttt{base-nts} $>$ \texttt{base} by $+0.077$ on moving\_direction at 4B.

% ------------------------------------------------------------
\paragraph{TVBench activation patching (Qwen3-VL, \emph{moving\_direction}).}
The patching suite replicates on TVBench for Qwen3-VL on \emph{moving\_direction}
($n{=}232$; Table~\ref{tab:tvbench-patch}). This run's shuffle draws its own
permutation, so its \texttt{shuf}/\texttt{shuf-corr} baselines are not comparable
to the per-task table above. \texttt{act\_layer0} leaves residuals of $-0.12$/$-0.10$, six times the
TempCompass value, consistent with direction answers depending more directly on
visual content order; \texttt{act\_ds} closes them at both scales
($+0.026$/$-0.009$ vs \texttt{base}, within one SE at $n{=}232$), and
\texttt{act\_early} adds nothing beyond it. The residual is larger here, and
the depth at which it closes is the same. Conv3d patching (\texttt{e1}) tracks \texttt{shuf}
($+0.013$/$-0.009$), confirming no independent Conv3d channel here either.
\texttt{shuf-corr} at 8B sits below \texttt{shuf} ($0.095$ vs $0.121$), within
one SE at this $n$; matched-permutation comparisons for shuffle contrasts are
in Appendix~\ref{app:seeds}.

\begin{table}[h]
\centering
\small
\caption{TVBench \emph{moving\_direction} activation patching, Qwen3-VL
($n{=}232$, seed 42, this run's own shuffle permutation).}
\label{tab:tvbench-patch}
\begin{tabular}{lcc}
\toprule
\textbf{Condition} & \textbf{4B} & \textbf{8B} \\
\midrule
\texttt{base}        & 0.461 & 0.397 \\
\texttt{shuf}        & 0.203 & 0.121 \\
\texttt{shuf-corr}   & 0.254 & 0.095 \\
\texttt{e1} (Conv3d) & 0.216 & 0.112 \\
\texttt{act\_layer0} & 0.341 & 0.297 \\
\texttt{act\_ds}     & 0.487 & 0.388 \\
\texttt{act\_early}  & 0.453 & 0.388 \\
\bottomrule
\end{tabular}
\end{table}

\section{Reversed-Label Scoring on TVBench}
\label{app:revgt}
% ------------------------------------------------------------

Full two-label scoring for the three TVBench tasks whose labels transform
deterministically under time reversal, across all reversal conditions
(main-text Table~\ref{tab:revgt} carries the \texttt{rev}/\texttt{rev-corr}
columns for \emph{moving\_direction} and \emph{scene\_transition}).
Reversed-label construction: \emph{moving\_direction} flips both motion axes
and the flipped option is always among the four candidates;
\emph{scene\_transition} swaps ``from A to B'' to ``from B to A'' (one item
unmapped and excluded); \emph{action\_sequence} swaps first and last, and both
candidates are actions that occur in the clip by construction~\citep{tvbench},
so the reversed label is the other candidate. Items whose reversed label is
not offered are excluded, not scored.

\begin{table}[h]
\centering
\small
\caption{TVBench two-label scoring, all reversal conditions. $E$: forward
reading, $F$: reversed reading; remainder to 1 is the distractor rate
(zero by construction on the two-option tasks).}
\label{tab:revgt-full}
\resizebox{\columnwidth}{!}{%
\begin{tabular}{llcccccc}
\toprule
 & & \multicolumn{2}{c}{\texttt{rev}} & \multicolumn{2}{c}{\texttt{rev-corr}} & \multicolumn{2}{c}{\texttt{rev-mvts}} \\
\cmidrule(lr){3-4}\cmidrule(lr){5-6}\cmidrule(lr){7-8}
\textbf{Task} & \textbf{Model} & $E$ & $F$ & $E$ & $F$ & $E$ & $F$ \\
\midrule
moving\_dir.\ ($n{=}232$) & Qwen 4B  & 0.065 & 0.629 & 0.056 & 0.608 & 0.168 & 0.431 \\
 & Qwen 8B  & 0.022 & 0.789 & 0.047 & 0.651 & 0.043 & 0.802 \\
 & Molmo 4B & 0.034 & 0.647 & 0.621 & 0.052 & 0.034 & 0.655 \\
 & Molmo 8B & 0.043 & 0.603 & 0.672 & 0.078 & 0.013 & 0.797 \\
\midrule
scene\_trans.\ ($n{=}184$) & Qwen 4B  & 0.223 & 0.777 & 0.310 & 0.690 & 0.326 & 0.674 \\
 & Qwen 8B  & 0.228 & 0.772 & 0.315 & 0.685 & 0.223 & 0.777 \\
 & Molmo 4B & 0.217 & 0.783 & 0.875 & 0.125 & 0.245 & 0.755 \\
 & Molmo 8B & 0.217 & 0.783 & 0.853 & 0.147 & 0.212 & 0.788 \\
\midrule
action\_seq.\ ($n{=}437$) & Qwen 4B  & 0.426 & 0.574 & 0.453 & 0.547 & 0.446 & 0.554 \\
 & Qwen 8B  & 0.452 & 0.548 & 0.489 & 0.511 & 0.468 & 0.532 \\
 & Molmo 4B & 0.382 & 0.618 & 0.824 & 0.176 & 0.375 & 0.625 \\
 & Molmo 8B & 0.364 & 0.636 & 0.824 & 0.176 & 0.371 & 0.629 \\
\bottomrule
\end{tabular}}
\end{table}

\subsection*{Why Qwen scores higher on reversed \emph{moving\_direction} clips than forward ones}
\label{app:revgt-asym}

Table~\ref{tab:revgt} shows Qwen at $E{=}0.47$/$0.38$ on the forward
\emph{moving\_direction} clips but $F{=}0.63$/$0.79$ on the coherent backwards
video, while Molmo goes the expected way ($0.70$/$0.69$ vs $0.65$/$0.60$). This
is not a property of the reversed labels. All 232 items offer the same four
options (up/down $\times$ left/right) with 58 items per class under either
reading, so a fixed answer preference scores exactly $0.25$ against both
labels; frame counts, prompt, timestamps, and RoPE are identical between
\texttt{base} and \texttt{rev}, and only the pixel order differs.

The asymmetry is a stimulus property meeting a readout property. The clips
are CLEVRER renders viewed from above: objects enter from the frame edge with
initial velocity and decelerate under friction, and ``up'' in the image is
motion away from the camera, so up-movers are typically objects entering from
the near edge whose motion is concentrated in the early frames, while
down-movers are already in scene and still moving at the end. Whole-frame
motion energy (mean absolute frame difference at 8\,fps) places $0.57$ of the
motion in the first half of the clip for up-GT items and $0.41$ for down-GT
items. Table~\ref{tab:revgt-asym} bins items by that front share. Qwen-8B's
forward accuracy falls from $0.46$ to $0.09$ as the motion moves to the front
of the clip while its reversed accuracy stays near $0.8$, and Molmo is flat:
Qwen-8B reads direction from the late frames, and reversal moves the motion
there. On forward up-movers it is at chance on both axes and answers ``down''
($47$ of the 58 lowest-late-motion clips, with chosen letters spread across
A--D, so the default is semantic rather than positional). Qwen-4B, which
samples five frames of these clips, shows neither the late-frame weighting nor
the downward default, so we do not describe it as a family property. The
reversed clip is therefore an easier stimulus for Qwen, and \texttt{rev} $F$
is the appropriate reference for \texttt{rev-corr} $F$ (conflict cost $0.02$/$0.14$),
which is how Table~\ref{tab:revgt} reads it. Whole-frame energy is a proxy for
the queried object's motion, so this is a scoping explanation, not a mechanism
claim.

\begin{table}[h]
\centering
\small
\caption{\emph{moving\_direction} accuracy by the share of whole-frame motion
energy in the first half of the clip. \texttt{base} $E$: forward clip against
the forward label; \texttt{rev} $F$: reversed clip against the reversed label.}
\label{tab:revgt-asym}
\resizebox{\columnwidth}{!}{%
\begin{tabular}{lccccccc}
\toprule
 & & \multicolumn{2}{c}{Qwen 8B} & \multicolumn{2}{c}{Qwen 4B} & \multicolumn{2}{c}{Molmo 8B} \\
\cmidrule(lr){3-4}\cmidrule(lr){5-6}\cmidrule(lr){7-8}
\textbf{Front share} & $n$ & \texttt{base} $E$ & \texttt{rev} $F$ & \texttt{base} $E$ & \texttt{rev} $F$ & \texttt{base} $E$ & \texttt{rev} $F$ \\
\midrule
$[0.0, 0.5)$ & 120 & 0.46 & 0.79 & 0.45 & 0.55 & 0.66 & 0.65 \\
$[0.5, 0.6)$ &  56 & 0.39 & 0.82 & 0.46 & 0.68 & 0.66 & 0.61 \\
$[0.6, 0.7)$ &  45 & 0.22 & 0.78 & 0.51 & 0.78 & 0.80 & 0.49 \\
$[0.7, 1.0]$ &  11 & 0.09 & 0.64 & 0.55 & 0.64 & 0.73 & 0.55 \\
\bottomrule
\end{tabular}}
\end{table}

% ------------------------------------------------------------
\section{Reversed-Clip Base Control}
\label{app:revclip}
% ------------------------------------------------------------

Full results for the reversed-content control of
Section~\ref{sec:behavior:controls}: overall and per-aspect, all seven
models (Table~\ref{tab:revclip-aspect}). The 95 \texttt{\{id\}\_reverse}
clips are valid forward videos evaluated under
\texttt{base}. Items are paired by (base video id, aspect, question stem);
four direction items are excluded because TempCompass rewrote the question
text itself under reversal (appear/disappear), and one video pair (13+3
items) is absent from the Molmo2-8B run, hence its $n{=}294$. \emph{fwd-read
(all)} is the fraction of all reverse-clip predictions matching the paired
original's GT; \emph{fwd-read (err)} restricts to errors.

\emph{rand} is the expected fwd-read (err) if errors picked uniformly among
the reverse item's wrong options, the random-error baseline. The five
Molmo2/Qwen3-VL models and PLM 3B sit well above it (overall fwd-read (err)
$0.59$ to $0.75$ against about $0.37$). PLM 1B sits at or below it on both
aspects ($0.403$ vs $0.394$ on attribute\_change, $0.236$ vs $0.376$ on
direction). The attribute\_change null is informative because PLM 1B scores
$0.875$ on the paired originals, so it does the task on forward content and
fails unsystematically on reversed content rather than substituting the
forward reading. On direction its base accuracy is $0.484$, so that aspect
has little power to reveal a prior and we do not interpret it (its reversal
drop CI also crosses zero, as does PLM 3B's). Whether the model lacks the
prior or lacks the comprehension of reversed content needed to express one,
this control cannot distinguish.

\begin{table}[h]
\centering
\small
\caption{Reversed-content control: \texttt{base} accuracy on the 95
\texttt{\_reverse} clips vs paired originals, overall and per aspect.
Paired bootstrap 95\% CIs on $\Delta$. \emph{fwd-read (err)}: fraction of
reverse-clip errors that are exactly the forward-content answer.
\emph{rand}: expected fwd-read (err) under uniformly random wrong-option
errors.}
\label{tab:revclip-aspect}
\resizebox{\columnwidth}{!}{%
\begin{tabular}{llcccccc}
\toprule
\textbf{Model} & \textbf{Aspect} & $n$ & orig & rev & $\Delta$ [CI] & fwd-read (err) & rand \\
\midrule
Molmo2 4B & attr.\ change & 144 & 0.965 & 0.674 & $-0.292$ $[-0.38,-0.22]$ & 0.830 & 0.379 \\
Molmo2 4B & direction     & 153 & 0.627 & 0.444 & $-0.183$ $[-0.30,-0.07]$ & 0.671 & 0.365 \\
Molmo2 4B & overall       & 297 & 0.791 & 0.556 & $-0.236$ $[-0.31,-0.16]$ & 0.727 & 0.370 \\
Molmo2 8B & attr.\ change & 144 & 0.972 & 0.701 & $-0.271$ $[-0.35,-0.19]$ & 0.860 & 0.384 \\
Molmo2 8B & direction     & 150 & 0.607 & 0.447 & $-0.160$ $[-0.27,-0.04]$ & 0.663 & 0.363 \\
Molmo2 8B & overall       & 294 & 0.786 & 0.571 & $-0.214$ $[-0.29,-0.14]$ & 0.730 & 0.370 \\
Molmo2 O-7B & attr.\ change & 144 & 0.965 & 0.653 & $-0.312$ $[-0.40,-0.23]$ & 0.740 & 0.373 \\
Molmo2 O-7B & direction   & 153 & 0.608 & 0.477 & $-0.131$ $[-0.26,-0.01]$ & 0.637 & 0.367 \\
Molmo2 O-7B & overall     & 297 & 0.781 & 0.562 & $-0.219$ $[-0.29,-0.14]$ & 0.677 & 0.369 \\
Qwen3-VL 4B & attr.\ change & 144 & 0.944 & 0.618 & $-0.326$ $[-0.42,-0.24]$ & 0.891 & 0.379 \\
Qwen3-VL 4B & direction   & 153 & 0.654 & 0.327 & $-0.327$ $[-0.43,-0.22]$ & 0.670 & 0.372 \\
Qwen3-VL 4B & overall     & 297 & 0.795 & 0.468 & $-0.327$ $[-0.40,-0.26]$ & 0.747 & 0.374 \\
Qwen3-VL 8B & attr.\ change & 144 & 0.972 & 0.674 & $-0.299$ $[-0.38,-0.22]$ & 0.936 & 0.387 \\
Qwen3-VL 8B & direction   & 153 & 0.608 & 0.444 & $-0.163$ $[-0.28,-0.05]$ & 0.647 & 0.357 \\
Qwen3-VL 8B & overall     & 297 & 0.785 & 0.556 & $-0.229$ $[-0.30,-0.16]$ & 0.750 & 0.367 \\
PLM 1B & attr.\ change & 144 & 0.875 & 0.500 & $-0.375$ $[-0.48,-0.26]$ & 0.403 & 0.394 \\
PLM 1B & direction     & 153 & 0.484 & 0.418 & $-0.065$ $[-0.20,+0.07]$ & 0.236 & 0.376 \\
PLM 1B & overall       & 297 & 0.673 & 0.458 & $-0.215$ $[-0.30,-0.13]$ & 0.311 & 0.384 \\
PLM 3B & attr.\ change & 144 & 0.938 & 0.625 & $-0.312$ $[-0.40,-0.22]$ & 0.796 & 0.383 \\
PLM 3B & direction     & 153 & 0.536 & 0.471 & $-0.065$ $[-0.19,+0.06]$ & 0.444 & 0.370 \\
PLM 3B & overall       & 297 & 0.731 & 0.545 & $-0.185$ $[-0.26,-0.11]$ & 0.585 & 0.375 \\
\bottomrule
\end{tabular}}
\end{table}

% ------------------------------------------------------------
\section{Held-Out Motion-Dependent Slice}
\label{app:heldout}
% ------------------------------------------------------------

The motion-dependent slice is outcome-selected: an item enters model $M$'s
slice when $M$ answers it correctly from the full video but not from a single
frame. The held-out check defines the slice with one family's outcomes and
evaluates the other family's $F{-}E$ on it at the same scale, so the evaluated
model plays no role in item selection. Own-slice rows reproduce main-text
Table~\ref{tab:reversal-tc} exactly. On \emph{direction} (not shown) held-out
magnitudes shrink mechanically because the held-out slice no longer
guarantees the evaluated model answers the item at base; the paper draws no
\emph{direction} channel claims.

\begin{table}[h]
\centering
\small
\caption{Held-out slice check, TempCompass \emph{attribute\_change}. Paired
bootstrap 95\% CIs.}
\label{tab:heldout}
\resizebox{\columnwidth}{!}{%
\begin{tabular}{lcccc}
\toprule
\textbf{Evaluated on held-out slice} & $n$ & $E$ & $F$ & $F{-}E$ [CI] \\
\midrule
Qwen 4B on Molmo-4B slice & 67 & 0.433 & 0.493 & $+0.060$ $[-0.18,+0.30]$ \\
Molmo 4B on Qwen-4B slice & 71 & 0.986 & 0.000 & $-0.986$ $[-1.00,-0.96]$ \\
Qwen 8B on Molmo-8B slice & 59 & 0.305 & 0.627 & $+0.322$ $[+0.08,+0.54]$ \\
Molmo 8B on Qwen-8B slice & 76 & 0.987 & 0.000 & $-0.987$ $[-1.00,-0.96]$ \\
\bottomrule
\end{tabular}}
\end{table}

% ------------------------------------------------------------
\section{Permutation-Seed Spreads}
\label{app:seeds}
% ------------------------------------------------------------

Key shuffled conditions were rerun with three permutation seeds. The seeded
runner derives one RNG per (seed, task, video, question) and shares a single
permutation across paired shuffle conditions per video, so paired contrasts
(e.g.\ pixel-shuffle vs RoPE-shuffle) compare identical stimuli; the original
runs drew independent permutations per condition. Observed spreads: sd
$\leq 0.0118$ for micro-averaged scores (TempCompass overall $n{=}1580$,
TVBench $n{=}2205$) and $\leq 0.0124$ for the Molmo TVBench nine-task macro
average. The unified noise rule in the main text (differences below $2$\,sd
of the noisiest same-benchmark shuffled condition) follows from these
spreads. Spreads vary by condition and scale without a consistent channel
pattern: \texttt{rop-shuf} shows the smallest spreads in the 4B-era runs
but not at 8B (Qwen 8B TVBench \texttt{rop-shuf} sd $0.0057$ against
\texttt{shuf}'s $0.0019$; Molmo 8B \texttt{rop-shuf} $0.0085$ against
\texttt{shuf}'s $0.0045$). On matched stimuli the Molmo TVBench
\texttt{rop-shuf} minus \texttt{shuf} gap is positive in all seven draws
across the three scales ($+0.005$/$+0.016$ at 4B, up to $+0.034$ at 8B),
\texttt{rop-shuf} consistently the slightly less costly.

\begin{table}[h]
\centering
\small
\caption{Seed spreads (overall accuracy; mean $\pm$ sd over three seeds). TVBench rows are micro-averaged except those marked macro.}
\label{tab:seeds}
\resizebox{\columnwidth}{!}{%
\begin{tabular}{llcc}
\toprule
\textbf{Run} & \textbf{Condition} & \textbf{mean} & \textbf{sd} \\
\midrule
Qwen 4B TempCompass & \texttt{shuf}              & 0.5656 & 0.0071 \\
Qwen 4B TempCompass & \texttt{shuf-corr} (RoPE)  & 0.6110 & 0.0004 \\
Qwen 4B TempCompass & \texttt{shuf-corr} (+DS)   & 0.6171 & 0.0061 \\
Qwen 4B TempCompass & \texttt{rop-shuf}          & 0.6709 & 0.0006 \\
Qwen 4B TVBench     & \texttt{shuf}              & 0.4080 & 0.0054 \\
Qwen 4B TVBench     & \texttt{shuf-corr} (RoPE)  & 0.4624 & 0.0043 \\
Qwen 4B TVBench     & \texttt{rop-shuf}          & 0.5339 & 0.0016 \\
Molmo 4B TVBench (macro) & \texttt{shuf}         & 0.3845 & 0.0124 \\
Molmo 4B TVBench (macro) & \texttt{rop-shuf}     & 0.4021 & 0.0065 \\
Qwen 8B TempCompass & \texttt{shuf}              & 0.5789 & 0.0118 \\
Qwen 8B TVBench     & \texttt{shuf}              & 0.3755 & 0.0019 \\
Qwen 8B TVBench     & \texttt{shuf-corr} (RoPE)  & 0.4299 & 0.0046 \\
Qwen 8B TVBench     & \texttt{rop-shuf}          & 0.4942 & 0.0057 \\
Molmo 8B TVBench (macro) & \texttt{shuf}         & 0.3789 & 0.0045 \\
Molmo 8B TVBench (macro) & \texttt{rop-shuf}     & 0.4079 & 0.0085 \\
Molmo O-7B TVBench (macro) & \texttt{shuf}       & 0.3623 & 0.0061 \\
Molmo O-7B TVBench (macro) & \texttt{rop-shuf}   & 0.3950 & 0.0016 \\
\bottomrule
\end{tabular}}
\end{table}

\end{document}